\documentclass[10pt,twocolumn,letterpaper]{article}

\usepackage{iccv}
\usepackage{times}
\usepackage{epsfig}
\usepackage{graphicx}
\usepackage{amsmath}
\usepackage{amssymb}
\usepackage{bm}

\usepackage{subfigure}
\usepackage{multirow}
\usepackage{array}
\usepackage{authblk}

\usepackage[pagebackref=true,breaklinks=true,letterpaper=true,colorlinks,bookmarks=false]{hyperref}

\iccvfinalcopy 

\ificcvfinal\fi

\makeatletter

\newcommand{\Rmnum}[1]{\expandafter\@slowromancap\romannumeral #1@}
\makeatother

\begin{document}

\title{StackGAN: Text to Photo-realistic Image Synthesis \\ with Stacked Generative Adversarial Networks}

\author[1]{Han Zhang} 
\author[2]{\;Tao Xu} 
\author[3]{\;Hongsheng Li} 
\author[4]{\\Shaoting Zhang}
\author[3]{\;Xiaogang  Wang}
\author[2]{\;Xiaolei Huang}
\author[1]{\;Dimitris Metaxas}

\affil[ ]{$^{1}$Rutgers University \;  $^{2}$Lehigh University \; $^{3}$The Chinese University of Hong Kong \;  $^{4}$Baidu Research}
\affil[ ]{\tt\small {\{han.zhang,~dnm\}@cs.rutgers.edu, \{tax313,~xih206\}@lehigh.edu}}
\affil[ ]{\tt\small {\{hsli,~xgwang\}@ee.cuhk.edu.hk, zhangshaoting@baidu.com}}
\renewcommand\Authands{, }  

%

\maketitle

\begin{abstract}

Synthesizing high-quality images from text descriptions is a challenging problem in computer vision and has many practical applications. Samples generated by existing text-to-image approaches can roughly reflect the meaning of the given descriptions, but they fail to contain necessary details and vivid object parts. In this paper, we propose Stacked Generative Adversarial Networks (StackGAN) to generate 256$\times$256 photo-realistic images conditioned on text descriptions. We decompose the hard problem into more manageable sub-problems through a sketch-refinement process. The Stage-\Rmnum{1} GAN sketches the primitive shape and colors of the object based on the given text description, yielding Stage-\Rmnum{1} low-resolution images. The Stage-\Rmnum{2} GAN takes Stage-\Rmnum{1} results and text descriptions as inputs, and generates high-resolution images with photo-realistic details. It is able to rectify defects in Stage-\Rmnum{1} results and add compelling details with the refinement process. To improve the diversity of the synthesized images and stabilize the training of the conditional-GAN, we introduce a novel Conditioning Augmentation technique that encourages smoothness in the latent conditioning manifold. Extensive experiments and comparisons with state-of-the-arts on benchmark datasets demonstrate that the proposed method achieves significant improvements on generating photo-realistic images conditioned on text descriptions.

\end{abstract}


\vspace{-8pt}
\section{Introduction}
\vspace{-5pt}

Generating photo-realistic images from text is an important problem and has tremendous applications, including photo-editing, computer-aided design, \etc. Recently, Generative Adversarial Networks (GAN)~\cite{goodfellow2014generative, DentonCSF15, Radford15} have shown promising results in synthesizing real-world images. Conditioned on given text descriptions, conditional-GANs~\cite{reed2016generative, reed2016learning} are able to generate images that are highly related to the text meanings.

\begin{figure}[bt]
\begin{center}
	\includegraphics[width=0.95\linewidth]{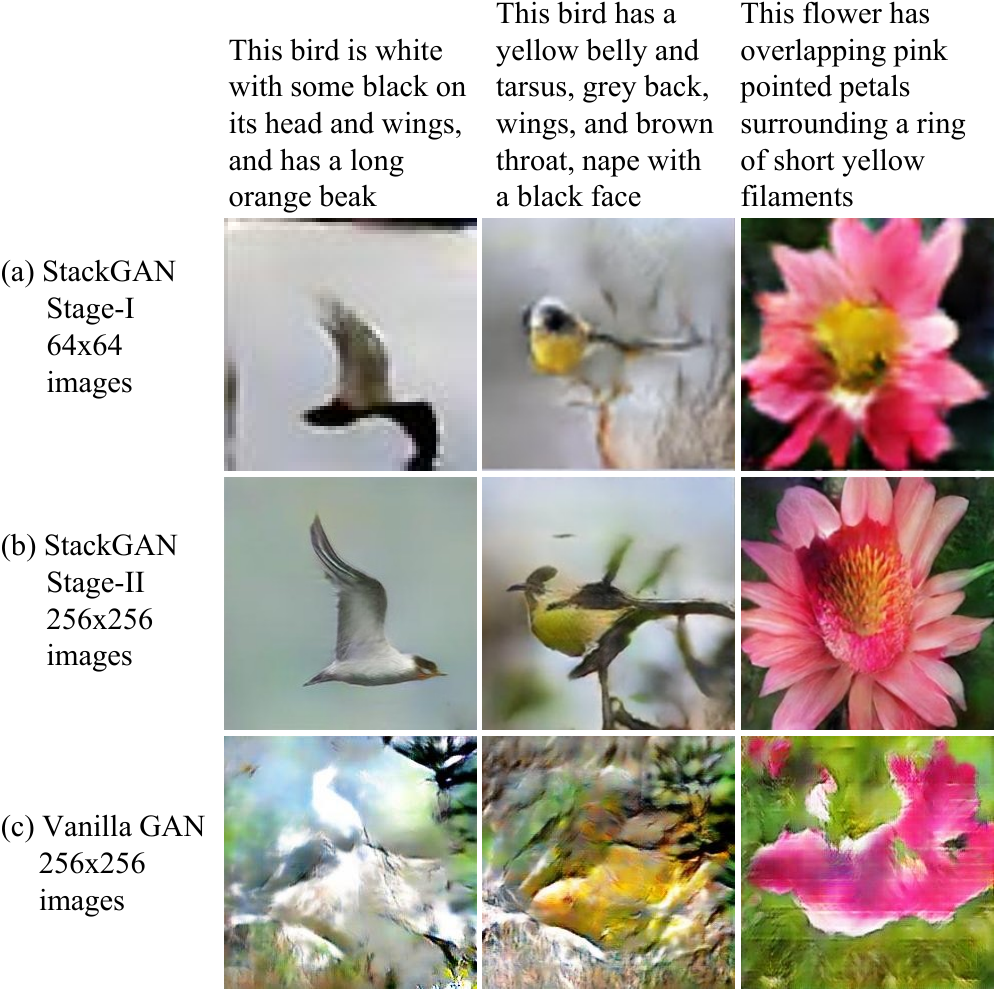}
\end{center}
\vspace{-8pt}
   \caption{
    Comparison of the proposed StackGAN and a vanilla one-stage GAN for generating 256$\times$256 images.    (a) Given text descriptions, Stage-\Rmnum{1} of StackGAN sketches rough shapes and basic colors of objects, yielding low-resolution images.    (b) Stage-\Rmnum{2} of StackGAN takes Stage-\Rmnum{1} results and text descriptions as inputs, and generates high-resolution images with photo-realistic details.    (c) Results by a vanilla 256$\times$256 GAN which simply adds more upsampling layers to state-of-the-art GAN-INT-CLS~\cite{reed2016generative}. It is unable to generate any plausible images of 256$\times$256 resolution.}
    \label{fig:big_examples}
\vspace{-8pt}
\end{figure}

However, it is very difficult to train GAN to generate high-resolution photo-realistic images from text descriptions. Simply adding more upsampling layers in state-of-the-art GAN models for generating high-resolution (\eg, 256$\times$256) images generally results in training instability and produces nonsensical outputs (see Figure~\ref{fig:big_examples}(c)). The main difficulty for generating high-resolution images by GANs is that supports of natural image distribution and implied model distribution may not overlap in high dimensional pixel space~\cite{Casper2016, ArjovskyB17}. This problem is more severe as the image resolution increases. Reed \etal only succeeded in generating plausible 64$\times$64 images conditioned on text descriptions~\cite{reed2016generative}, which usually lack details and vivid object parts, \eg, beaks and eyes of birds. Moreover, they were unable to synthesize higher resolution (\eg, 128$\times$128) images without providing additional annotations of objects~\cite{reed2016learning}.

In analogy to how human painters draw, we decompose the problem of text to photo-realistic image synthesis into two more tractable sub-problems with Stacked Generative Adversarial Networks (StackGAN). Low-resolution images are first generated by our Stage-\Rmnum{1} GAN (see Figure~\ref{fig:big_examples}(a)). On the top of our Stage-\Rmnum{1} GAN, we stack Stage-\Rmnum{2} GAN to generate realistic high-resolution (\eg, 256$\times$256) images conditioned on Stage-\Rmnum{1} results and text descriptions (see Figure~\ref{fig:big_examples}(b)). By conditioning on the Stage-\Rmnum{1} result and the text again, Stage-\Rmnum{2} GAN learns to capture the text information that is omitted by Stage-\Rmnum{1} GAN and draws more details for the object. The support of model distribution generated from a roughly aligned low-resolution image has better probability of intersecting with the support of image distribution. This is the underlying reason why Stage-\Rmnum{2} GAN is able to generate better high-resolution images.

In addition, for the text-to-image generation task, the limited number of training text-image pairs often results in sparsity in the text conditioning manifold and such sparsity makes it difficult to train GAN. Thus, we propose a novel Conditioning Augmentation technique to encourage smoothness in the latent conditioning manifold. It allows small random perturbations in the conditioning manifold and increases the diversity of synthesized images. 

The contribution of the proposed method is threefold: 
(1) We propose a novel Stacked Generative Adversarial Networks for synthesizing \emph{photo-realistic} images from text descriptions. It decomposes the difficult problem of generating high-resolution images into more manageable subproblems and significantly improve the state of the art. The StackGAN for the first time generates images of 256$\times$256 resolution with photo-realistic details from text descriptions. 
(2) A new Conditioning Augmentation technique is proposed to stabilize the conditional GAN training and also improves the diversity of the generated samples. 
(3) Extensive qualitative and quantitative experiments demonstrate the effectiveness of the overall model design as well as the effects of individual components, which provide useful information for designing future conditional GAN models. 
Our code is available at {\href{https://github.com/hanzhanggit/StackGAN}{https://github.com/hanzhanggit/StackGAN.}}


\vspace{-2pt}
\section{Related Work}
 \vspace{-5pt}

Generative image modeling is a fundamental problem in computer vision. There has been remarkable progress in this direction with the emergence of deep learning techniques. Variational Autoencoders (VAE)~\cite{KingmaW14, RezendeMW14} formulated the problem with probabilistic graphical models whose goal was to maximize the lower bound of data likelihood. Autoregressive models (\eg, PixelRNN)~\cite{OordKK16} that utilized neural networks to model the conditional distribution of the pixel space have also generated appealing synthetic images. Recently, Generative Adversarial Networks (GAN)~\cite{goodfellow2014generative} have shown promising performance for generating sharper images. But training instability makes it hard for GAN models to generate high-resolution (\eg, 256$\times$256) images. Several techniques~\cite{Radford15, Salimans2016, MetzICLR17, ArjovskyB17, CheLJBL16} have been proposed to stabilize the training process and generate compelling results. An energy-based GAN~\cite{Zhao2016} has also been proposed for more stable training behavior.  

Built upon these generative models, conditional image generation has also been studied. Most methods utilized simple conditioning variables such as attributes or class labels~\cite{YanYSL16, Oord16, ChenDHSSA16, Odena2016}. There is also work conditioned on images to generate images, including photo editing~\cite{Brock2016, ZhuKSE16}, domain transfer~\cite{Taigmaniclr17, pix2pix2017} and super-resolution~\cite{Casper2016, Christian2016}. However, super-resolution methods~\cite{Casper2016, Christian2016} can only add limited details to low-resolution images and can not correct large defects as our proposed StackGAN does. Recently, several methods have been developed to generate images from unstructured text. Mansimov \etal~\cite{MansimovPBS15} built an AlignDRAW model by learning to estimate alignment between text and the generating canvas.  Reed \etal~\cite{reed2016iclr17} used conditional PixelCNN to generate images using the text descriptions and object location constraints. Nguyen \etal~\cite{NguyenYBDC17} used an approximate Langevin sampling approach to generate images conditioned on text. However, their sampling approach requires an inefficient iterative optimization process. With conditional GAN, Reed \etal~\cite{reed2016generative} successfully generated plausible 64$\times$64 images for birds and flowers based on text descriptions. Their follow-up work~\cite{reed2016learning} was able to generate 128$\times$128 images by utilizing additional annotations on object part locations.

Besides using a single GAN for generating images, there is also work~\cite{WangG16, DentonCSF15,huang2016sgan} that utilized a series of GANs for image generation. Wang \etal~\cite{WangG16} factorized the indoor scene generation process into structure generation and style generation with the proposed $S^2$-GAN. In contrast, the second stage of our StackGAN aims to complete object details and correct defects of Stage-\Rmnum{1} results based on text descriptions. Denton \etal~\cite{DentonCSF15} built a series of GANs within a Laplacian pyramid framework. At each level of the pyramid, a residual image was generated conditioned on the image of the previous stage and then added back to the input image to produce the input for the next stage. Concurrent to our work, Huang \etal~\cite{huang2016sgan} also showed that they can generate better images by stacking several GANs to reconstruct the multi-level representations of a pre-trained discriminative model. However, they only succeeded in generating 32$\times$32 images, while our method utilizes a simpler architecture to generate 256$\times$256 images with photo-realistic details and sixty-four times more pixels.

\begin{figure*}[tb]
\begin{center}
\includegraphics[width=0.85\linewidth]{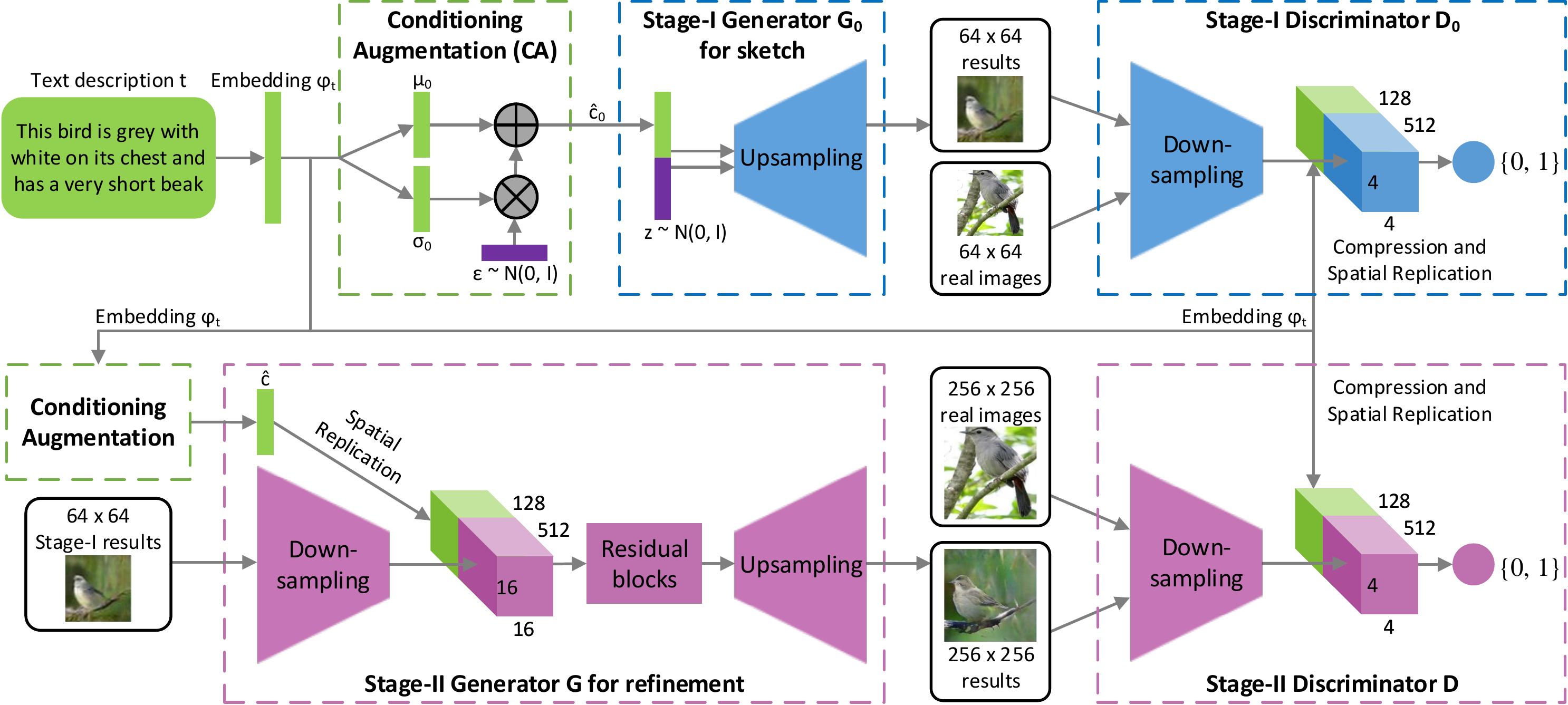}
   \caption{The architecture of the proposed StackGAN. The Stage-\Rmnum{1} generator draws a low-resolution image by sketching rough shape and basic colors of the object from the given text and painting the background from a random noise vector. Conditioned on Stage-\Rmnum{1} results, the Stage-\Rmnum{2} generator corrects defects and adds compelling details into Stage-\Rmnum{1} results, yielding a more realistic high-resolution image.
   }
	\vspace{-12pt}
\end{center}
\label{fig:framework}
\end{figure*}

\vspace{-2pt}
\section{Stacked Generative Adversarial Networks}
\vspace{-5pt}

To generate high-resolution images with photo-realistic details, we propose a simple yet effective Stacked Generative Adversarial Networks. It decomposes the text-to-image generative process into two stages (see Figure~\ref{fig:framework}).  
\vspace{-5pt}
\begin{enumerate}
\item [-] \textbf{Stage-\Rmnum{1} GAN:} it sketches the primitive shape and basic colors of the object conditioned on the given text description, and draws the background layout from a random noise vector, yielding a low-resolution image. 
\vspace{-10pt}
\item [-] \textbf{Stage-\Rmnum{2} GAN:} it corrects defects in the low-resolution image from Stage-I and completes details of the object by reading the text description again, producing a high-resolution photo-realistic image. 
\end{enumerate}
\vspace{-5pt}

\subsection{Preliminaries}
\vspace{-5pt}
Generative Adversarial Networks (GAN)~\cite{goodfellow2014generative} are composed of two models that are alternatively trained to compete with each other. The generator $G$ is optimized to reproduce the true data distribution $p_{data}$ by generating images that are difficult for the discriminator $D$ to differentiate from real images. Meanwhile, $D$ is optimized to distinguish real images and synthetic images generated by $G$. Overall, the training procedure is similar to a two-player min-max game with the following objective function, 
\begin{equation}\label{eq:GAN_ori}
\begin{aligned}
\min_{G} \max_{D} V(D,G) = \; & \mathbb{E}_{x \sim {p_{data}}} [\log D(x)] \; + \\
& \mathbb{E}_{z \sim {p_{z}}} [\log(1 - D(G(z)))],
\end{aligned}
\end{equation}
where $x$ is a real image from the true data distribution $p_{data}$, and $z$ is a noise vector sampled from distribution $p_{z}$ (\eg, uniform or Gaussian distribution). 

Conditional GAN~\cite{gauthier2015conditional, Mirza14} is an extension of GAN where both the generator and discriminator receive additional conditioning variables $c$, yielding $G(z,c)$ and $D(x,c)$. This formulation allows $G$ to generate images conditioned on variables $c$. 

\subsection{Conditioning Augmentation}
\vspace{-5pt}
As shown in Figure~\ref{fig:framework}, the text description $t$ is first encoded by an encoder, yielding a text embedding $\varphi_{t}$. In previous works~\cite{reed2016generative,reed2016learning}, the text embedding is nonlinearly transformed to generate conditioning latent variables as the input of the generator. However, latent space for the text embedding is usually high dimensional ($>100$ dimensions). With limited amount of data, it usually causes discontinuity in the latent data manifold, which is not desirable for learning the generator. To mitigate this problem, we introduce a Conditioning Augmentation technique to produce additional conditioning variables $\hat{c}$. In contrast to the fixed conditioning text variable $c$ in \cite{reed2016generative,reed2016learning}, we randomly sample the latent variables $\hat{c}$ from an independent Gaussian distribution $\mathcal{N}(\mu(\varphi_{t}), \Sigma(\varphi_{t}))$, where the mean $\mu(\varphi_{t})$ and diagonal covariance matrix $\Sigma(\varphi_{t})$ are functions of the text embedding $\varphi_{t}$. The proposed Conditioning Augmentation yields more training pairs given a small number of image-text pairs, and thus encourages robustness to small perturbations along the conditioning manifold. To further enforce the smoothness over the conditioning manifold and avoid overfitting~\cite{Doersch16, LarsenSLW16}, we add the following regularization term to the objective of the generator during training, 
\begin{equation}\label{eq:GaussianKL}
\begin{aligned}
D_{KL}(\mathcal{N}(\mu(\varphi_{t}), \Sigma(\varphi_{t})) \, || \, \mathcal{N}(0, I)),
\end{aligned}
\end{equation}
which is the Kullback-Leibler divergence (KL divergence) between the standard Gaussian distribution and the conditioning Gaussian distribution. 
The randomness introduced in the Conditioning Augmentation is beneficial for modeling text to image translation as the same sentence usually corresponds to objects with various poses and appearances.

\subsection{Stage-\Rmnum{1} GAN}
\vspace{-5pt}

Instead of directly generating a high-resolution image conditioned on the text description, we simplify the task to first generate a low-resolution image with our Stage-\Rmnum{1} GAN, which focuses on drawing only rough shape and correct colors for the object. 

Let $\varphi_{t}$ be the text embedding of the given description, which is generated by a pre-trained encoder~\cite{reed2016cvpr} in this paper. The Gaussian conditioning variables $\hat{c}_0$ for text embedding are sampled from $\mathcal{N}(\mu_0(\varphi_{t}), \Sigma_0(\varphi_{t}))$ to capture the meaning of $\varphi_{t}$ with variations. 
Conditioned on $\hat{c}_0$ and random variable $z$, Stage-\Rmnum{1} GAN trains the discriminator $D_0$ and the generator $G_0$ by alternatively maximizing $\mathcal{L}_{D_0}$ in Eq.~(\ref{eq:D_0}) and minimizing $\mathcal{L}_{G_0}$ in Eq.~(\ref{eq:G_0}), 
\begin{equation}\label{eq:D_0}
\begin{aligned}
\mathcal{L}_{D_0} =  &\; \mathbb{E}_{(I_{0},t) \sim {p_{data}}} [\log D_{0}(I_{0}, \varphi_{t})] \; + \\
&\; \mathbb{E}_{z \sim {p_{z}}, t \sim p_{data}} [\log(1 - D_{0}(G_{0}(z,\hat{c}_0), \varphi_{t}))],
\end{aligned}
\end{equation}
\begin{equation}\label{eq:G_0}
\begin{aligned}
\mathcal{L}_{G_0} = &\; \mathbb{E}_{z \sim {p_{z}}, t \sim p_{data}} [\log(1 - D_{0}(G_{0}(z,\hat{c}_0), \varphi_{t}))] \; + \\
                    &\; \lambda D_{KL}(\mathcal{N}(\mu_0(\varphi_{t}), \Sigma_0(\varphi_{t})) \, || \, \mathcal{N}(0, I)), 
\end{aligned}
\end{equation}
where the real image $I_0$ and the text description $t$ are from the true data distribution $p_{data}$. 
$z$ is a noise vector randomly sampled from a given distribution $p_z$ (Gaussian distribution in this paper). 
$\lambda$ is a regularization parameter that balances the two terms in Eq.~(\ref{eq:G_0}).
We set $\lambda=1$ for all our experiments. 
Using the reparameterization trick introduced in~\cite{KingmaW14}, both $\mu_0(\varphi_{t})$ and $\Sigma_0(\varphi_{t})$ are learned jointly with the rest of the network.  

\textbf{Model Architecture. }
For the generator $G_0$, to obtain text conditioning variable $\hat{c}_0$, the text embedding $\varphi_{t}$ is first fed into a fully connected layer to generate $\mu_0$ and $\sigma_0$ ($\sigma_0$ are the values in the diagonal of $\Sigma_0$) for the Gaussian distribution $\mathcal{N}(\mu_0(\varphi_{t}), \Sigma_0(\varphi_{t}))$. $\hat{c}_0$ are then sampled from the Gaussian distribution.
Our $N_g$ dimensional conditioning vector $\hat{c_0}$ is computed by $\hat{c_0} = \mu_0 + \sigma_0 \odot \epsilon$ (where $\odot$ is the element-wise multiplication, $\epsilon  \sim \mathcal{N}(0, I)$). 
Then, $\hat{c_0}$ is concatenated with a $N_z$ dimensional noise vector to generate a $W_{0} \times H_{0}$ image by a series of up-sampling blocks.  

For the discriminator $D_0$, the text embedding $\varphi_{t}$ is first compressed to $N_d$ dimensions using a fully-connected layer and then spatially replicated to form a $M_d \times M_d \times N_d$ tensor. 
Meanwhile, the image is fed through a series of down-sampling blocks until it has $M_d \times M_d$ spatial dimension. 
Then, the image filter map is concatenated along the channel dimension with the text tensor. 
The resulting tensor is further fed to a 1$\times$1 convolutional layer to jointly learn features across the image and the text. 
Finally, a fully-connected layer with one node is used to produce the decision score. 

\subsection{Stage-\Rmnum{2} GAN}
\vspace{-5pt}

Low-resolution images generated by Stage-\Rmnum{1} GAN usually lack vivid object parts and might contain shape distortions. Some details in the text might also be omitted in the first stage, which is vital for generating photo-realistic images.  
Our Stage-\Rmnum{2} GAN is built upon Stage-\Rmnum{1} GAN results to generate high-resolution images. 
It is conditioned on low-resolution images and also the text embedding again to correct defects in Stage-\Rmnum{1} results. The Stage-\Rmnum{2} GAN completes previously ignored text information to generate more photo-realistic details.

Conditioning on the low-resolution result $s_0 = G_0(z, \hat{c}_0)$ and Gaussian latent variables $\hat{c}$, the discriminator $D$ and generator $G$ in Stage-\Rmnum{2} GAN are trained by alternatively maximizing $\mathcal{L}_{D}$ in Eq.~(\ref{eq:D}) and minimizing $\mathcal{L}_{G}$ in Eq.~(\ref{eq:G}), 
\begin{equation}\label{eq:D}
\begin{aligned}
\mathcal{L}_{D} =  &\; \mathbb{E}_{(I,t) \sim {p_{data}}} [\log D(I, \varphi_{t})] \; + \\
&\; \mathbb{E}_{s_0 \sim {p_{G_0}}, t \sim p_{data}} [\log(1 - D(G(s_0, \hat{c}), \varphi_{t}))],
\end{aligned}
\end{equation}
\begin{equation}\label{eq:G}
\begin{aligned}
\mathcal{L}_{G} = &\; \mathbb{E}_{s_0 \sim {p_{G_0}}, t \sim p_{data}} [\log(1 - D(G(s_0, \hat{c}), \varphi_{t}))] \; + \\
                    &\; \lambda D_{KL}(\mathcal{N}(\mu(\varphi_{t}), \Sigma(\varphi_{t})) \, || \, \mathcal{N}(0, I)), 
\end{aligned}
\end{equation}
Different from the original GAN formulation, the random noise $z$ is not used in this stage with the assumption that the randomness has already been preserved by $s_0$. 
Gaussian conditioning variables $\hat{c}$ used in this stage and $\hat{c}_0$ used in Stage-\Rmnum{1} GAN share the same pre-trained text encoder, generating the same text embedding $\varphi_{t}$. 
However, Stage-I and Stage-II Conditioning Augmentation have different fully connected layers for generating different means and standard deviations. In this way, Stage-\Rmnum{2} GAN learns to capture useful information in the text embedding that is omitted by Stage-\Rmnum{1} GAN.

\textbf{Model Architecture. }
We design Stage-\Rmnum{2} generator as an encoder-decoder network with residual blocks~\cite{HeZRS15}. 
Similar to the previous stage, the text embedding $\varphi_{t}$ is used to generate the $N_g$ dimensional text conditioning vector $\hat{c}$, which is spatially replicated to form a $M_{g} \times M_{g} \times N_g$ tensor. 
Meanwhile, the Stage-\Rmnum{1} result $s_0$ generated by Stage-\Rmnum{1} GAN is fed into several down-sampling blocks (\ie, encoder) until it has a spatial size of $M_{g} \times M_{g}$. The image features and the text features are concatenated along the channel dimension. 
The encoded image features coupled with text features are fed into several residual blocks, which are designed to learn multi-modal representations across image and text features. 
Finally, a series of up-sampling layers (\ie, decoder) are used to generate a $W \times H$ high-resolution image. Such a generator is able to help rectify defects in the input image while add more details to generate the realistic high-resolution image.

For the discriminator, its structure is similar to that of Stage-\Rmnum{1} discriminator with only extra down-sampling blocks since the image size is larger in this stage.  
To explicitly enforce GAN to learn better alignment between the image and the conditioning text, rather than using the vanilla discriminator, we adopt the matching-aware discriminator proposed by Reed \etal~\cite{reed2016generative} for both stages. 
During training, the discriminator takes real images and their corresponding text descriptions as positive sample pairs, whereas negative sample pairs consist of two groups. 
The first is real images with mismatched text embeddings, while the second is synthetic images with their corresponding text embeddings. 

\vspace{-2pt}
\subsection{Implementation details}
\vspace{-5pt}

The up-sampling blocks consist of the nearest-neighbor upsampling followed by a 3$\times$3 stride 1 convolution. Batch normalization~\cite{IoffeS15} and ReLU activation are applied after every convolution except the last one.  The residual blocks consist of  3$\times$3 stride 1 convolutions, Batch normalization and ReLU. Two residual blocks are used in 128$\times$128 StackGAN models while four are used in 256$\times$256 models. The down-sampling blocks consist of 4$\times$4 stride 2 convolutions, Batch normalization and LeakyReLU, except that the first one does not have Batch normalization.

By default, $N_g=128$,  $N_z=100$, $M_g=16$,  $M_d=4$, $N_d=128$,  $W_{0}=H_{0} = 64$ and $W=H=256$. 
For training, we first iteratively train $D_0$ and $G_0$ of Stage-\Rmnum{1} GAN for 600 epochs by fixing Stage-\Rmnum{2} GAN.
Then we iteratively train $D$ and $G$ of Stage-\Rmnum{2} GAN for another 600 epochs by fixing Stage-\Rmnum{1} GAN. 
All networks are trained using ADAM solver with batch size 64 and an initial learning rate of 0.0002.  
The learning rate is decayed to $1/2$ of its previous value every 100 epochs.

\section{Experiments}
\vspace{-5pt}

To validate our method, we conduct extensive quantitative and qualitative evaluations. Two state-of-the-art methods on text-to-image synthesis, GAN-INT-CLS \cite{reed2016generative} and GAWWN \cite{reed2016learning}, are compared. Results by the two compared methods are generated using the code released by their authors. In addition, we design several baseline models to investigate the overall design and important components of our proposed StackGAN. For the first baseline, we directly train Stage-\Rmnum{1} GAN for generating 64$\times$64 and 256$\times$256 images to investigate whether the proposed stacked structure and Conditioning Augmentation are beneficial. Then we modify our StackGAN to generate 128$\times$128 and 256$\times$256 images to investigate whether larger images by our method result in higher image quality. We also investigate whether inputting text at both stages of StackGAN is useful.

\vspace{-2pt}
\subsection{Datasets and evaluation metrics} 
\vspace{-5pt}

CUB~\cite{WahCUB_200_2011} contains 200 bird species with 11,788 images. Since 80\% of birds in this dataset have object-image size ratios of less than 0.5~\cite{WahCUB_200_2011}, as a pre-processing step, we crop all images to ensure that bounding boxes of birds have greater-than-0.75 object-image size ratios. Oxford-102~\cite{Nilsback08} contains 8,189 images of flowers from 102 different categories. To show the generalization capability of our approach, a more challenging dataset, MS COCO~\cite{LinMBHPRDZ14} is also utilized for evaluation. Different from CUB and Oxford-102, the MS COCO dataset contains images with multiple objects and various backgrounds. It has a training set with 80k images and a validation set with 40k images. Each image in COCO has 5 descriptions, while 10 descriptions are provided by \cite{reed2016cvpr} for every image in CUB and Oxford-102 datasets. Following the experimental setup in \cite{reed2016generative}, we directly use the training and validation sets provided by COCO, meanwhile we split CUB and Oxford-102 into class-disjoint training and test sets. 

\textbf{Evaluation metrics. }
It is difficult to evaluate the performance of generative models (\eg, GAN). We choose a recently proposed numerical assessment approach ``inception score''~\cite{Salimans2016} for quantitative evaluation,
\begin{equation}\label{eq:Inception_score}
\begin{aligned}
I = \exp(\mathbb{E}_{\bm{x}} D_{KL} (p(y|\bm{x}) \,||\, p(y))),
\end{aligned}
\end{equation}
where $\bm{x}$ denotes one generated sample, and $y$ is the label predicted by the Inception model~\cite{Szegedy2016}. The intuition behind this metric is that good models should generate diverse but meaningful images. Therefore, the KL divergence between the marginal distribution $p(y)$ and the conditional distribution $p(y|\bm{x})$ should be large. In our experiments, we directly use the pre-trained Inception model for COCO dataset. For fine-grained datasets, CUB and Oxford-102, we fine-tune an Inception model for each of them. As suggested in~\cite{Salimans2016}, we evaluate this metric on a large number of samples (\ie, 30k randomly selected samples) for each model.

Although the inception score has shown to well correlate with human perception on visual quality of samples~\cite{Salimans2016}, it cannot reflect whether the generated images are well conditioned on the given text descriptions. Therefore, we also conduct human evaluation. We randomly select 50 text descriptions for each class of CUB and Oxford-102 test sets. For COCO dataset, 4k text descriptions are randomly selected from its validation set. For each sentence, 5 images are generated by each model. Given the same text descriptions, 10 users (not including any of the authors) are asked to rank the results by different methods. The average ranks by human users are calculated to evaluate all compared methods.

\begin{table}[bt]
\begin{center}
\scriptsize
\begin{tabular}{|c|l|c|c|c|}
\hline
    Metric &Dataset &GAN-INT-CLS &GAWWN &Our StackGAN\\
\hline
\multirow{3}{3.2em}{Inception score}  
    &CUB    &2.88~$\pm$~.04 &3.62~$\pm$~.07 &{\bf 3.70~$\pm$~.04}   \\
\cline{2-5}
    &Oxford &2.66~$\pm$~.03 &/              &{\bf 3.20~$\pm$~.01}  \\
\cline{2-5}
    &COCO   &7.88~$\pm$~.07 &/              &{\bf 8.45~$\pm$~.03}  \\
\hline
\multirow{3}{3.2em}{Human rank}  
    &CUB    &2.81~$\pm$~.03 &1.99~$\pm$~.04  &{\bf 1.37~$\pm$~.02}  \\
\cline{2-5}
    &Oxford &1.87~$\pm$~.03 &/               &{\bf 1.13~$\pm$~.03}   \\
\cline{2-5}
    &COCO   &1.89~$\pm$~.04 &/               &{\bf 1.11~$\pm$~.03}  \\
\hline
\end{tabular}
\end{center}
\vspace{-5pt}
    \caption{Inception scores and average human ranks of our StackGAN, GAWWN~\cite{reed2016learning}, and GAN-INT-CLS~\cite{reed2016generative} on CUB, Oxford-102, and MS-COCO datasets.}
\vspace{-15pt}
\label{tab:cmp_previous}
\end{table}
\begin{figure*}[tb]
\begin{center}
\includegraphics[width=1.0\linewidth]{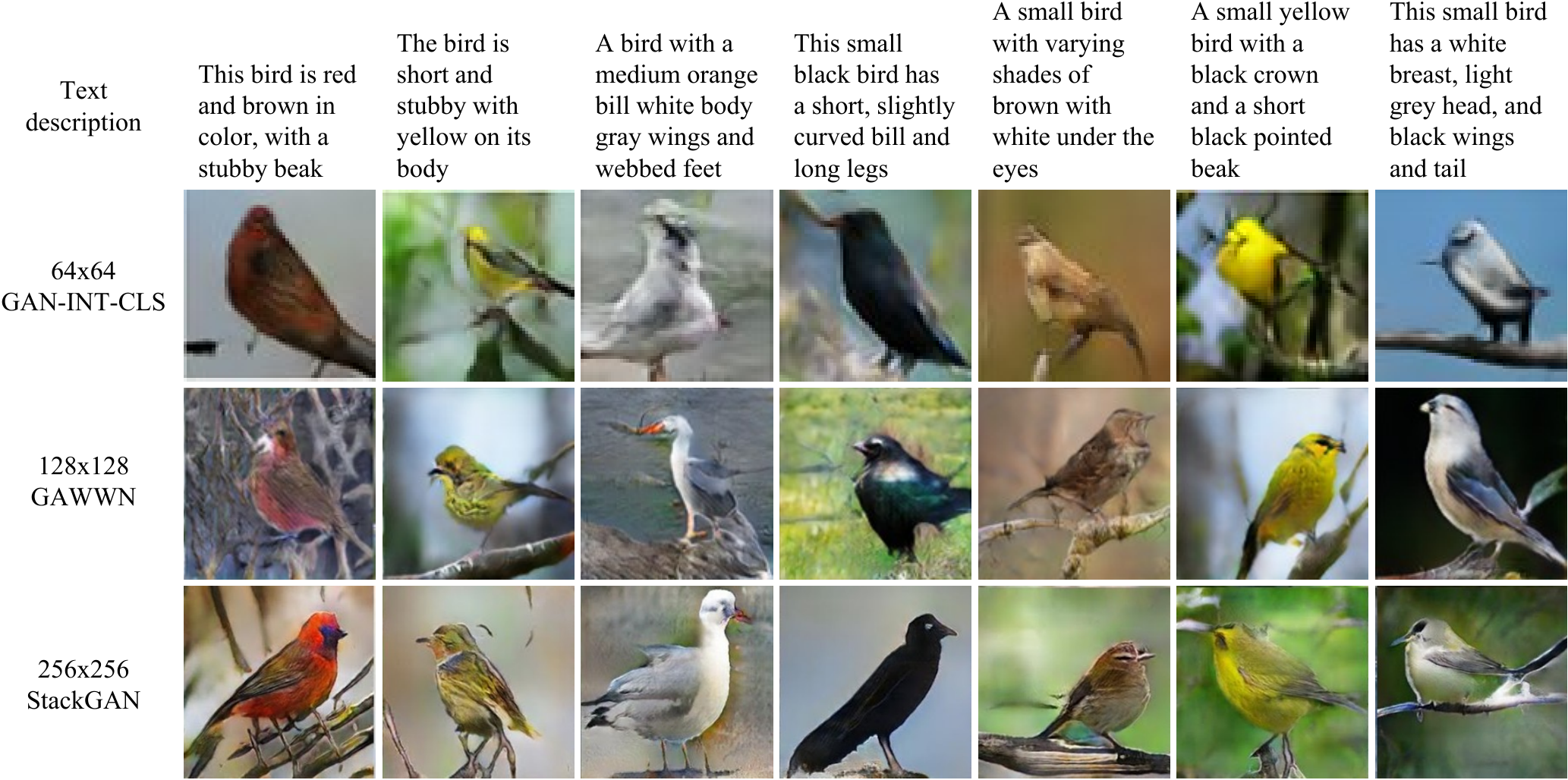}
\end{center}
\vspace{-8pt}
    \caption{Example results by our StackGAN, GAWWN~\cite{reed2016learning}, and GAN-INT-CLS~\cite{reed2016generative} conditioned on text descriptions from CUB test set.}
\vspace{-2pt}
\label{fig:cmp_previous}
\end{figure*}
\begin{figure*}[tb]
\begin{center}
\includegraphics[width=1.0\linewidth]{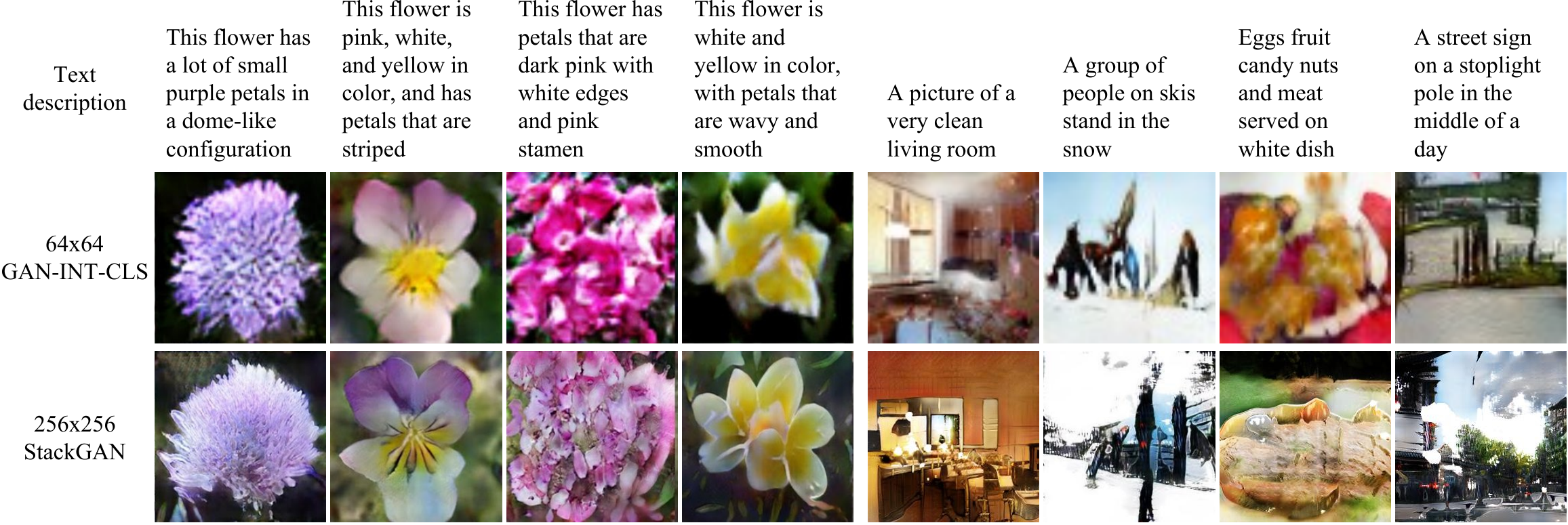}
\end{center}
\vspace{-8pt}
    \caption{Example results by our StackGAN and GAN-INT-CLS~\cite{reed2016generative} conditioned on text descriptions from Oxford-102 test set (leftmost four columns) and COCO validation set (rightmost four columns).}
 \vspace{-8pt}
\label{fig:cmp_previous_flower}
\end{figure*}
\begin{figure*}[bt]
\begin{center}
	\includegraphics[width=1.0\linewidth]{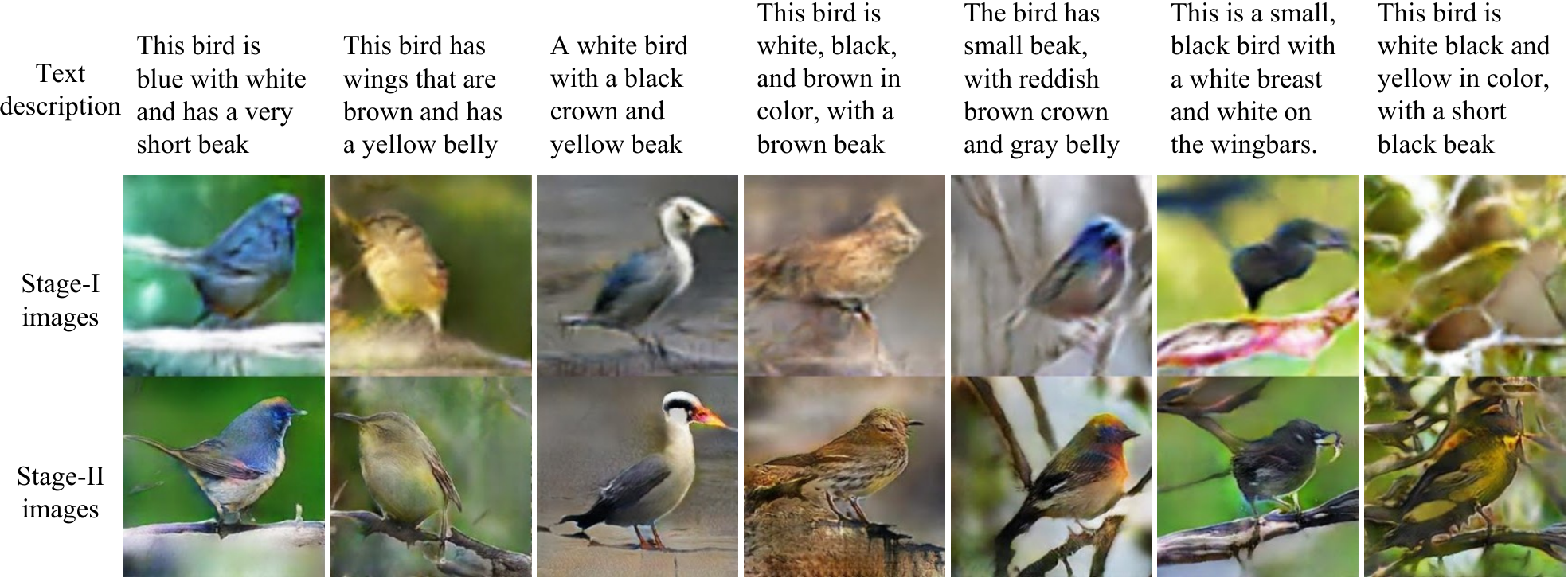}
\end{center}
\vspace{-8pt}
   \caption{Samples generated by our StackGAN from unseen texts in CUB test set. 
   Each column lists the text description, images generated from the text by Stage-\Rmnum{1} and Stage-\Rmnum{2} of StackGAN.}
\vspace{-10pt}
\label{fig:lr2hr}
\end{figure*}
\begin{figure}[bt]
\begin{center}
	\includegraphics[width=1.0\linewidth]{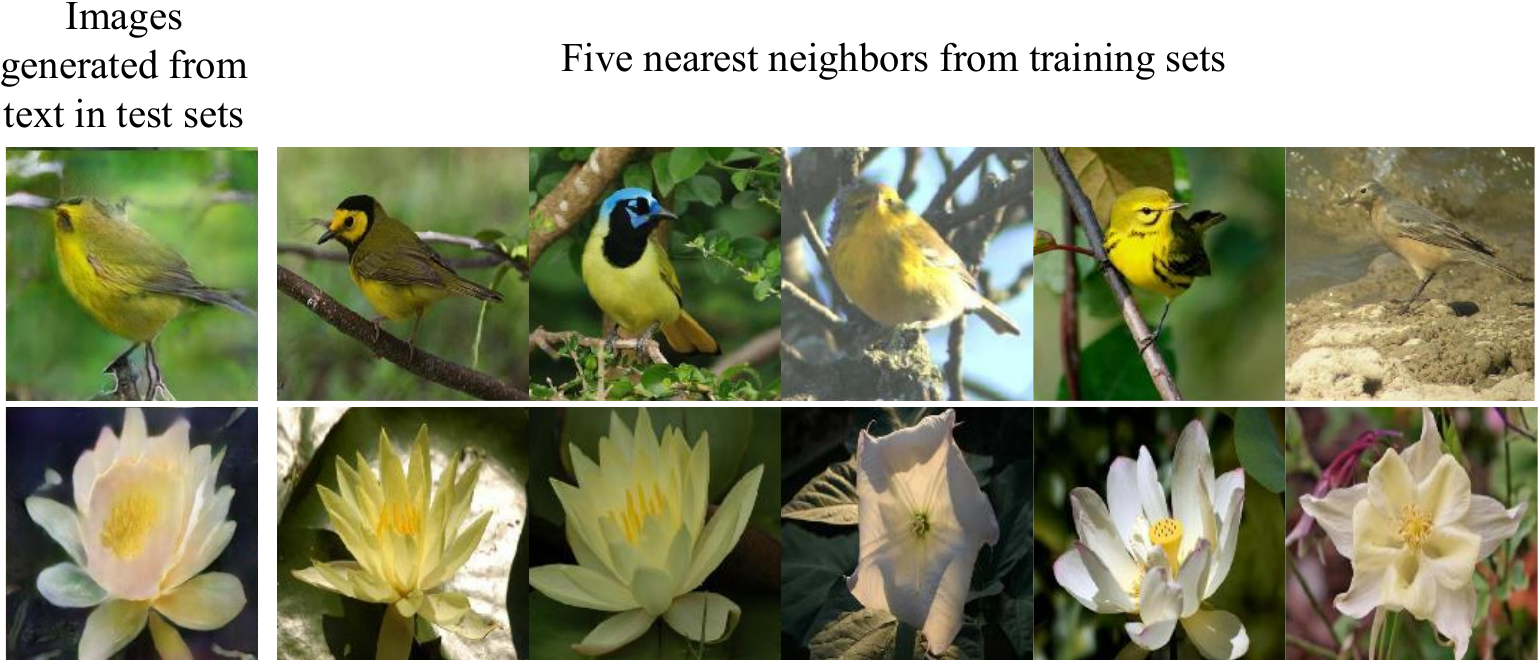}
\end{center}
\vspace{-8pt}
   \caption{For generated images (column 1), retrieving their nearest training images (columns 2-6) by utilizing Stage-II discriminator $D$ to extract visual features. The $L2$ distances between features are calculated for nearest-neighbor retrieval.}
\vspace{-10pt}
\label{fig:NNs}
\end{figure}

\vspace{-2pt}
\subsection{Quantitative and qualitative results}
\vspace{-5pt}

We compare our results with the state-of-the-art text-to-image methods~\cite{reed2016learning, reed2016generative} on CUB, Oxford-102 and COCO datasets. The inception scores and average human ranks for our proposed StackGAN and compared methods are reported in Table~\ref{tab:cmp_previous}. Representative examples are compared in Figure~\ref{fig:cmp_previous} and Figure~\ref{fig:cmp_previous_flower}.

Our StackGAN achieves the best inception score and average human rank on all three datasets. Compared with GAN-INT-CLS~\cite{reed2016generative}, StackGAN achieves 28.47\% improvement in terms of inception score on CUB dataset (from 2.88 to 3.70), and 20.30\% improvement on Oxford-102 (from 2.66 to 3.20). The better average human rank of our StackGAN also indicates our proposed method is able to generate more realistic samples conditioned on text descriptions.

As shown in Figure~\ref{fig:cmp_previous}, the 64$\times$64 samples generated by GAN-INT-CLS~\cite{reed2016generative} can only reflect the general shape and color of the birds. Their results lack vivid parts (\eg, beak and legs) and convincing details in most cases, which make them neither realistic enough nor have sufficiently high resolution. By using additional conditioning variables on location constraints, GAWWN~\cite{reed2016learning} obtains a better inception score on CUB dataset, which is still slightly lower than ours. It generates higher resolution images with more details than GAN-INT-CLS, as shown in Figure~\ref{fig:cmp_previous}. However, as mentioned by its authors, GAWWN fails to generate any plausible images when it is only conditioned on text descriptions~\cite{reed2016learning}. In comparison, our StackGAN can generate 256$\times$256 photo-realistic images from only text descriptions.

Figure~\ref{fig:lr2hr} illustrates some examples of the Stage-I and Stage-II images generated by our StackGAN. As shown in the first row of Figure \ref{fig:lr2hr}, in most cases, Stage-\Rmnum{1} GAN is able to draw rough shapes and colors of objects given text descriptions. However, Stage-\Rmnum{1} images are usually blurry with various defects and missing details, especially for foreground objects. As shown in the second row, Stage-\Rmnum{2} GAN generates 4$\times$ higher resolution images with more convincing details to better reflect corresponding text descriptions. For cases where Stage-\Rmnum{1} GAN has generated plausible shapes and colors, Stage-\Rmnum{2} GAN completes the details. For instance, in the $1$st column of Figure \ref{fig:lr2hr}, with a satisfactory Stage-\Rmnum{1} result, Stage-\Rmnum{2} GAN focuses on drawing the short beak and white color described in the text as well as details for the tail and legs. In all other examples, different degrees of details are added to Stage-\Rmnum{2} images. In many other cases, Stage-\Rmnum{2} GAN is able to correct the defects of Stage-\Rmnum{1} results by processing the text description again. For example, while the Stage-\Rmnum{1} image in the $5$th column has a blue crown rather than the reddish brown crown described in the text, the defect is corrected by Stage-\Rmnum{2} GAN. In some extreme cases (\eg, the $7$th column of Figure \ref{fig:lr2hr}), even when Stage-\Rmnum{1} GAN fails to draw a plausible shape, Stage-\Rmnum{2} GAN is able to generate reasonable objects. We also observe that StackGAN has the ability to transfer background from Stage-\Rmnum{1} images and fine-tune them to be more realistic with higher resolution at Stage-\Rmnum{2}.

Importantly, the StackGAN does not achieve good results by simply memorizing training samples but by capturing the complex underlying language-image relations. We extract visual features from our generated images and all training images by the Stage-II discriminator $D$ of our StackGAN. For each generated image, its nearest neighbors from the training set can be retrieved. By visually inspecting the retrieved images (see Figure~\ref{fig:NNs}), we can conclude that the generated images have some similar characteristics with the training samples but are essentially different.

\begin{figure}[bt]
\begin{center}
	\includegraphics[width=1.0\linewidth]{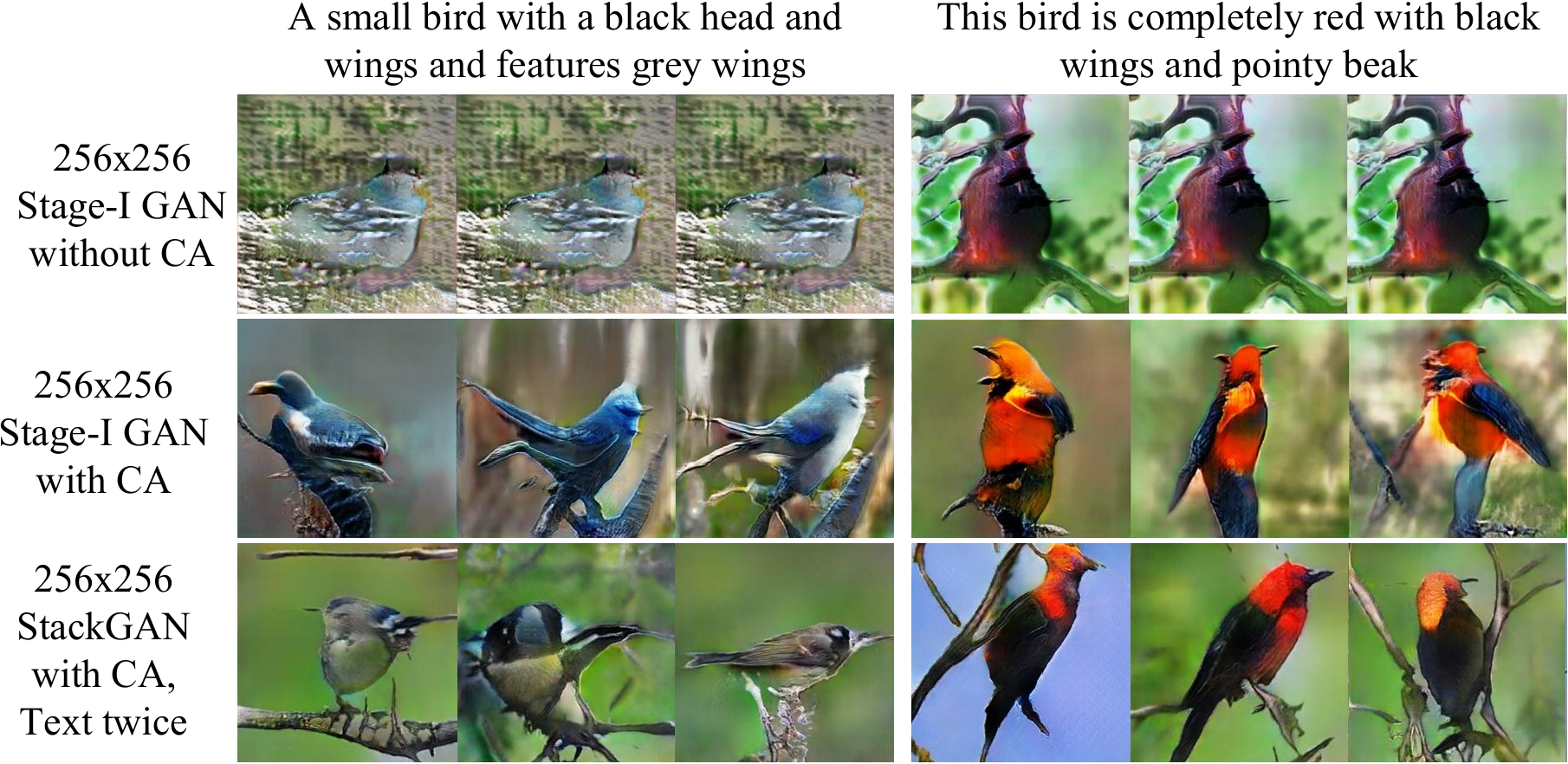}
\end{center}
\vspace{-8pt}
   \caption{Conditioning Augmentation (CA) helps stabilize the training of conditional GAN and improves the diversity of the generated samples. 
   (Row 1) without CA, Stage-\Rmnum{1} GAN fails to generate plausible 256$\times$256 samples. Although different noise vector $z$ is used for each column, the generated samples collapse to be the same for each input text description. 
   (Row 2-3) with CA but fixing the noise vectors $z$, methods are still able to generate birds with different poses and viewpoints.}
\vspace{-3pt}
\label{fig:Gaussian}
\end{figure}

\vspace{-2pt}
\subsection{Component analysis}
\vspace{-5pt}

In this subsection, we analyze different components of StackGAN on CUB dataset with our baseline models. 
The inception scores for those baselines are reported in Table~\ref{tab:inception_score}.

\textbf{The design of StackGAN.}　
As shown in the first four rows of Table~\ref{tab:inception_score}, if Stage-I GAN is directly used to generate images, the inception scores decrease significantly. Such performance drop can be well illustrated by results in Figure~\ref{fig:Gaussian}. As shown in the first row of Figure~\ref{fig:Gaussian}, Stage-\Rmnum{1} GAN fails to generate any plausible 256$\times$256 samples without using Conditioning Augmentation (CA). Although Stage-\Rmnum{1} GAN with CA is able to generate more diverse 256$\times$256 samples, those samples are not as realistic as samples generated by StackGAN. It demonstrates the necessity of the proposed stacked structure. In addition, by decreasing the output resolution from 256$\times$256 to 128$\times$128, the inception score decreases from 3.70 to 3.35. 
Note that all images are scaled to 299 $\times$ 299 before calculating the inception score. Thus, if our StackGAN just increases the image size without adding more information, the inception score would remain the same for samples of different resolutions. Therefore, the decrease in inception score by 128$\times$128 StackGAN demonstrates that our 256$\times$256 StackGAN does add more details into the larger images. For the 256$\times$256 StackGAN, if the text is only input to Stage-I (denoted as ``no Text twice''), the inception score decreases from 3.70 to 3.45. It indicates that processing text descriptions again at Stage-II helps refine Stage-I results. The same conclusion can be drawn from the results of 128$\times$128 StackGAN models.

\begin{table}
\begin{center}
\scriptsize
\begin{tabular}{|c|c|c|c|}
\hline
Method &CA &Text twice  &Inception score \\
\hline
64$\times$64 Stage-\Rmnum{1} GAN  &no    &/     &2.66~$\pm$~.03   \\
                                  &yes   &/     &2.95~$\pm$~.02   \\
\hline
\multirow{2}{3cm}{\,\,\,\,\,\,\,\,256$\times$256 Stage-\Rmnum{1} GAN}
                                  &no    &/     &2.48~$\pm$~.00   \\
                                  &yes    &/     &3.02~$\pm$~.01   \\
\hline
\multirow{3}{3cm}{\,\,\,\,\,\,\,\,\,\,\,128$\times$128 StackGAN}
                                  &yes   &no    &3.13~$\pm$~.03  \\ 
                                  &no    &yes   &3.20~$\pm$~.03  \\
                                  &yes   &yes   &3.35~$\pm$~.02  \\
\hline
\multirow{3}{3cm}{\,\,\,\,\,\,\,\,\,\,\,256$\times$256 StackGAN}
                                  &yes   &no    &3.45~$\pm$~.02  \\ 
                                  &no    &yes   &3.31~$\pm$~.03  \\
                                  &yes   &yes   &3.70~$\pm$~.04  \\
\hline
\end{tabular}
\end{center}
\vspace{-5pt}
    \caption{Inception scores calculated with 30,000 samples generated by different baseline models of our StackGAN.}
\label{tab:inception_score}
\vspace{-10pt}
\end{table}
\begin{figure}[bt]
\begin{center}
	\includegraphics[width=0.9\linewidth]{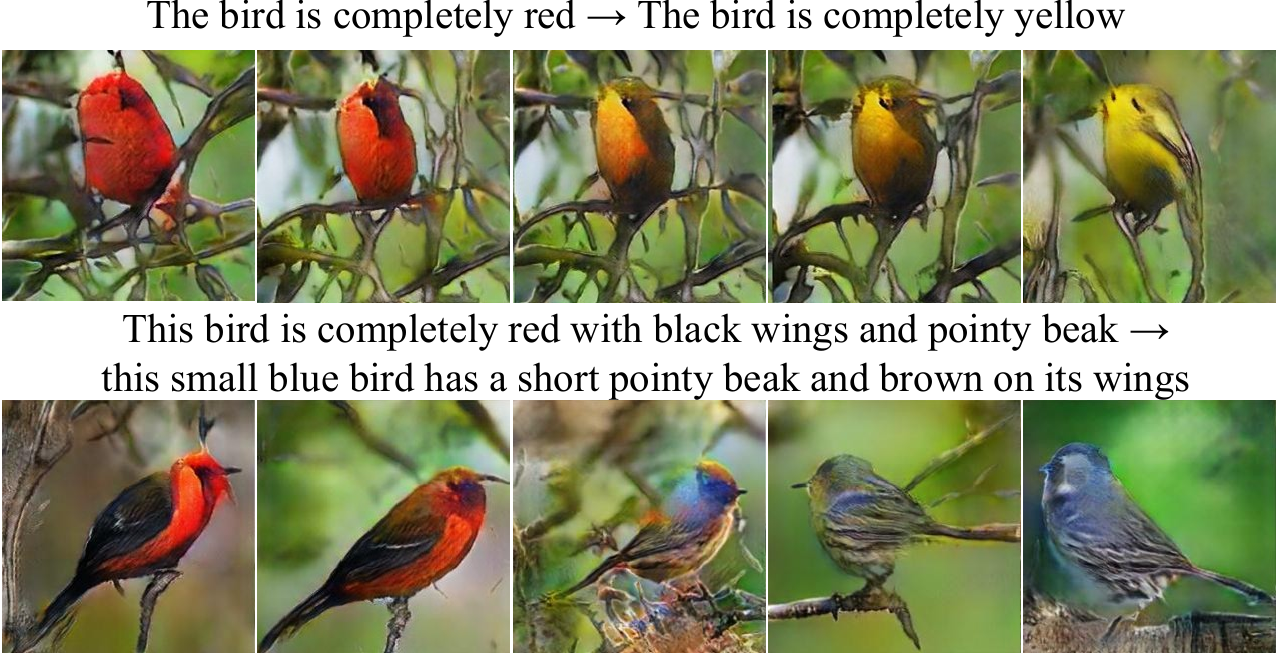}
\end{center}
\vspace{-8pt}
   \caption{(Left to right) Images generated by interpolating two sentence embeddings. Gradual appearance changes from the first sentence's meaning to that of the second sentence can be observed. The noise vector $z$ is fixed to be zeros for each row.}
\vspace{-10pt}
\label{fig:interpolate}
\end{figure}

\textbf{Conditioning Augmentation.} 
We also investigate the efficacy of the proposed Conditioning Augmentation (CA). By removing it from StackGAN 256$\times$256 (denoted as ``no CA'' in Table~\ref{tab:inception_score}), the inception score decreases from 3.70 to 3.31. Figure~\ref{fig:Gaussian} also shows that 256$\times$256 Stage-\Rmnum{1} GAN (and StackGAN) with CA can generate birds with different poses and viewpoints from the same text embedding. In contrast, without using CA, samples generated by 256$\times$256 Stage-\Rmnum{1} GAN collapse to nonsensical images due to the unstable training dynamics of GANs. Consequently, the proposed Conditioning Augmentation helps stabilize the conditional GAN training and improves the diversity of the generated samples because of its ability to encourage robustness to small perturbations along the latent manifold.

\textbf{Sentence embedding interpolation. }
To further demonstrate that our StackGAN learns a smooth latent data manifold, we use it to generate images from linearly interpolated sentence embeddings, as shown in Figure~\ref{fig:interpolate}. We fix the noise vector $z$, so the generated image is inferred from the given text description only. Images in the first row are generated by simple sentences made up by us. Those sentences contain only simple color descriptions. The results show that the generated images from interpolated embeddings can accurately reflect color changes and generate plausible bird shapes. The second row illustrates samples generated from more complex sentences, which contain more details on bird appearances. The generated images change their primary color from red to blue, and change the wing color from black to brown.

\vspace{-5pt}
\section{Conclusions}
\vspace{-5pt}

In this paper, we propose Stacked Generative Adversarial Networks (StackGAN) with Conditioning Augmentation for synthesizing \emph{photo-realistic} images. The proposed method decomposes the text-to-image synthesis to a novel sketch-refinement process. Stage-\Rmnum{1} GAN sketches the object following basic color and shape constraints from given text descriptions. Stage-\Rmnum{2} GAN corrects the defects in Stage-\Rmnum{1} results and adds more details, yielding higher resolution images with better image quality. Extensive quantitative and qualitative results demonstrate the effectiveness of our proposed method. Compared to existing text-to-image generative models, our method generates higher resolution images (\eg, 256$\times$256) with more photo-realistic details and diversity. 

{\small
\bibliographystyle{ieee}
\bibliography{paperbib}

\begin{thebibliography}{10}\itemsep=-1pt

\bibitem{ArjovskyB17}
M.~Arjovsky and L.~Bottou.
\newblock Towards principled methods for training generative adversarial
  networks.
\newblock In {\em {ICLR}}, 2017.

\bibitem{Brock2016}
A.~Brock, T.~Lim, J.~M. Ritchie, and N.~Weston.
\newblock Neural photo editing with introspective adversarial networks.
\newblock In {\em {ICLR}}, 2017.

\bibitem{CheLJBL16}
T.~Che, Y.~Li, A.~P. Jacob, Y.~Bengio, and W.~Li.
\newblock Mode regularized generative adversarial networks.
\newblock In {\em {ICLR}}, 2017.

\bibitem{ChenDHSSA16}
X.~Chen, Y.~Duan, R.~Houthooft, J.~Schulman, I.~Sutskever, and P.~Abbeel.
\newblock Infogan: Interpretable representation learning by information
  maximizing generative adversarial nets.
\newblock In {\em {NIPS}}, 2016.

\bibitem{DentonCSF15}
E.~L. Denton, S.~Chintala, A.~Szlam, and R.~Fergus.
\newblock Deep generative image models using a laplacian pyramid of adversarial
  networks.
\newblock In {\em {NIPS}}, 2015.

\bibitem{Doersch16}
C.~Doersch.
\newblock Tutorial on variational autoencoders.
\newblock {\em arXiv:1606.05908}, 2016.

\bibitem{gauthier2015conditional}
J.~Gauthier.
\newblock Conditional generative adversarial networks for convolutional face
  generation.
\newblock {\em Technical report}, 2015.

\bibitem{goodfellow2014generative}
I.~J. Goodfellow, J.~Pouget{-}Abadie, M.~Mirza, B.~Xu, D.~Warde{-}Farley,
  S.~Ozair, A.~C. Courville, and Y.~Bengio.
\newblock Generative adversarial nets.
\newblock In {\em {NIPS}}, 2014.

\bibitem{HeZRS15}
K.~He, X.~Zhang, S.~Ren, and J.~Sun.
\newblock Deep residual learning for image recognition.
\newblock In {\em {CVPR}}, 2016.

\bibitem{huang2016sgan}
X.~Huang, Y.~Li, O.~Poursaeed, J.~Hopcroft, and S.~Belongie.
\newblock Stacked generative adversarial networks.
\newblock In {\em {CVPR}}, 2017.

\bibitem{IoffeS15}
S.~Ioffe and C.~Szegedy.
\newblock Batch normalization: Accelerating deep network training by reducing
  internal covariate shift.
\newblock In {\em {ICML}}, 2015.

\bibitem{pix2pix2017}
P.~Isola, J.-Y. Zhu, T.~Zhou, and A.~A. Efros.
\newblock Image-to-image translation with conditional adversarial networks.
\newblock In {\em {CVPR}}, 2017.

\bibitem{KingmaW14}
D.~P. Kingma and M.~Welling.
\newblock Auto-encoding variational bayes.
\newblock In {\em {ICLR}}, 2014.

\bibitem{LarsenSLW16}
A.~B.~L. Larsen, S.~K. S{\o}nderby, H.~Larochelle, and O.~Winther.
\newblock Autoencoding beyond pixels using a learned similarity metric.
\newblock In {\em {ICML}}, 2016.

\bibitem{Christian2016}
C.~Ledig, L.~Theis, F.~Huszar, J.~Caballero, A.~Aitken, A.~Tejani, J.~Totz,
  Z.~Wang, and W.~Shi.
\newblock Photo-realistic single image super-resolution using a generative
  adversarial network.
\newblock In {\em {CVPR}}, 2017.

\bibitem{LinMBHPRDZ14}
T.-Y. Lin, M.~Maire, S.~Belongie, J.~Hays, P.~Perona, D.~Ramanan, P.~Dollár,
  and C.~L. Zitnick.
\newblock Microsoft coco: Common objects in context.
\newblock In {\em {ECCV}}, 2014.

\bibitem{MansimovPBS15}
E.~Mansimov, E.~Parisotto, L.~J. Ba, and R.~Salakhutdinov.
\newblock Generating images from captions with attention.
\newblock In {\em {ICLR}}, 2016.

\bibitem{MetzICLR17}
L.~Metz, B.~Poole, D.~Pfau, and J.~Sohl{-}Dickstein.
\newblock Unrolled generative adversarial networks.
\newblock In {\em {ICLR}}, 2017.

\bibitem{Mirza14}
M.~Mirza and S.~Osindero.
\newblock Conditional generative adversarial nets.
\newblock {\em arXiv:1411.1784}, 2014.

\bibitem{NguyenYBDC17}
A.~Nguyen, J.~Yosinski, Y.~Bengio, A.~Dosovitskiy, and J.~Clune.
\newblock Plug {\&} play generative networks: Conditional iterative generation
  of images in latent space.
\newblock In {\em {CVPR}}, 2017.

\bibitem{Nilsback08}
M.-E. Nilsback and A.~Zisserman.
\newblock Automated flower classification over a large number of classes.
\newblock In {\em ICCVGIP}, 2008.

\bibitem{Odena2016}
A.~Odena, C.~Olah, and J.~Shlens.
\newblock Conditional image synthesis with auxiliary classifier gans.
\newblock In {\em {ICML}}, 2017.

\bibitem{Radford15}
A.~Radford, L.~Metz, and S.~Chintala.
\newblock Unsupervised representation learning with deep convolutional
  generative adversarial networks.
\newblock In {\em {ICLR}}, 2016.

\bibitem{reed2016learning}
S.~Reed, Z.~Akata, S.~Mohan, S.~Tenka, B.~Schiele, and H.~Lee.
\newblock Learning what and where to draw.
\newblock In {\em {NIPS}}, 2016.

\bibitem{reed2016cvpr}
S.~Reed, Z.~Akata, B.~Schiele, and H.~Lee.
\newblock Learning deep representations of fine-grained visual descriptions.
\newblock In {\em CVPR}, 2016.

\bibitem{reed2016generative}
S.~Reed, Z.~Akata, X.~Yan, L.~Logeswaran, B.~Schiele, and H.~Lee.
\newblock Generative adversarial text-to-image synthesis.
\newblock In {\em {ICML}}, 2016.

\bibitem{reed2016iclr17}
S.~Reed, A.~van~den Oord, N.~Kalchbrenner, V.~Bapst, M.~Botvinick, and
  N.~de~Freitas.
\newblock Generating interpretable images with controllable structure.
\newblock {\em Technical report}, 2016.

\bibitem{RezendeMW14}
D.~J. Rezende, S.~Mohamed, and D.~Wierstra.
\newblock Stochastic backpropagation and approximate inference in deep
  generative models.
\newblock In {\em {ICML}}, 2014.

\bibitem{Salimans2016}
T.~Salimans, I.~J. Goodfellow, W.~Zaremba, V.~Cheung, A.~Radford, and X.~Chen.
\newblock Improved techniques for training gans.
\newblock In {\em {NIPS}}, 2016.

\bibitem{Szegedy2016}
C.~Szegedy, V.~Vanhoucke, S.~Ioffe, J.~Shlens, and Z.~Wojna.
\newblock Rethinking the inception architecture for computer vision.
\newblock In {\em {CVPR}}, 2016.

\bibitem{Casper2016}
C.~K. Sønderby, J.~Caballero, L.~Theis, W.~Shi, and F.~Huszar.
\newblock Amortised map inference for image super-resolution.
\newblock In {\em {ICLR}}, 2017.

\bibitem{Taigmaniclr17}
Y.~Taigman, A.~Polyak, and L.~Wolf.
\newblock Unsupervised cross-domain image generation.
\newblock In {\em {ICLR}}, 2017.

\bibitem{OordKK16}
A.~van~den Oord, N.~Kalchbrenner, and K.~Kavukcuoglu.
\newblock Pixel recurrent neural networks.
\newblock In {\em {ICML}}, 2016.

\bibitem{Oord16}
A.~van~den Oord, N.~Kalchbrenner, O.~Vinyals, L.~Espeholt, A.~Graves, and
  K.~Kavukcuoglu.
\newblock Conditional image generation with pixelcnn decoders.
\newblock In {\em {NIPS}}, 2016.

\bibitem{WahCUB_200_2011}
C.~Wah, S.~Branson, P.~Welinder, P.~Perona, and S.~Belongie.
\newblock {The Caltech-UCSD Birds-200-2011 Dataset}.
\newblock Technical Report CNS-TR-2011-001, California Institute of Technology,
  2011.

\bibitem{WangG16}
X.~Wang and A.~Gupta.
\newblock Generative image modeling using style and structure adversarial
  networks.
\newblock In {\em {ECCV}}, 2016.

\bibitem{YanYSL16}
X.~Yan, J.~Yang, K.~Sohn, and H.~Lee.
\newblock Attribute2image: Conditional image generation from visual attributes.
\newblock In {\em {ECCV}}, 2016.

\bibitem{Zhao2016}
J.~Zhao, M.~Mathieu, and Y.~LeCun.
\newblock Energy-based generative adversarial network.
\newblock In {\em {ICLR}}, 2017.

\bibitem{ZhuKSE16}
J.~Zhu, P.~Kr{\"{a}}henb{\"{u}}hl, E.~Shechtman, and A.~A. Efros.
\newblock Generative visual manipulation on the natural image manifold.
\newblock In {\em {ECCV}}, 2016.

\end{thebibliography}
}

\clearpage
\onecolumn
\centerline{\Large \bf Supplementary Materials}
\section*{More Results of Birds and Flowers}
\textbf{Additional Results on CUB Dataset}
\vspace{+5pt}

\includegraphics[width=0.95\linewidth]{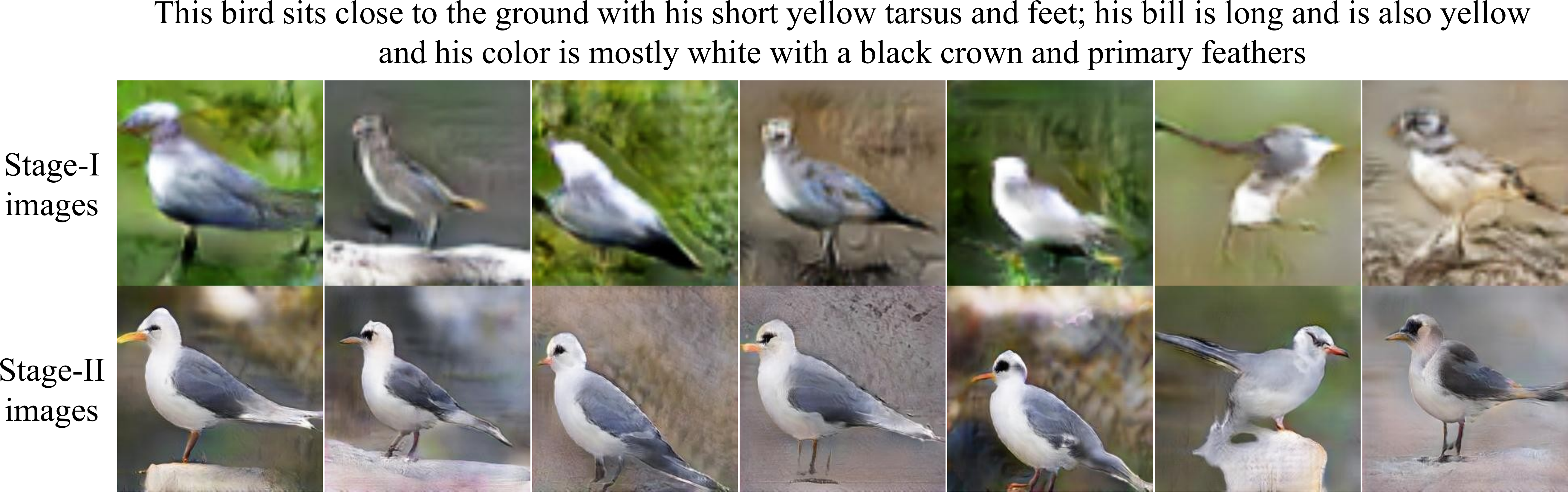}
\vspace{+5pt}

\includegraphics[width=0.95\linewidth]{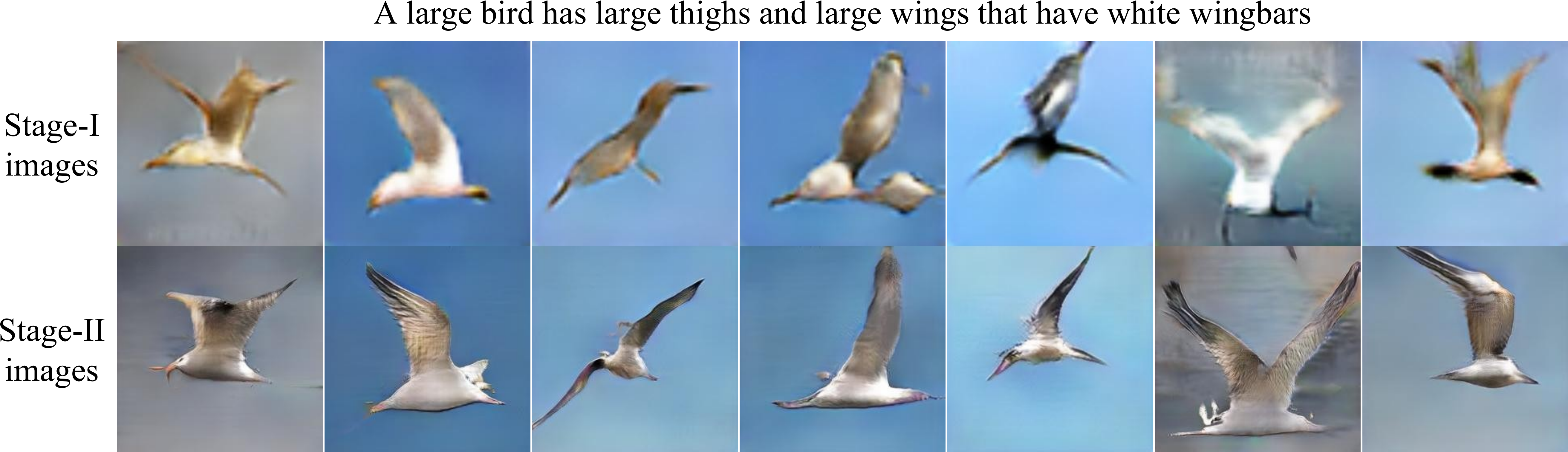}
\vspace{+5pt}

\includegraphics[width=0.95\linewidth]{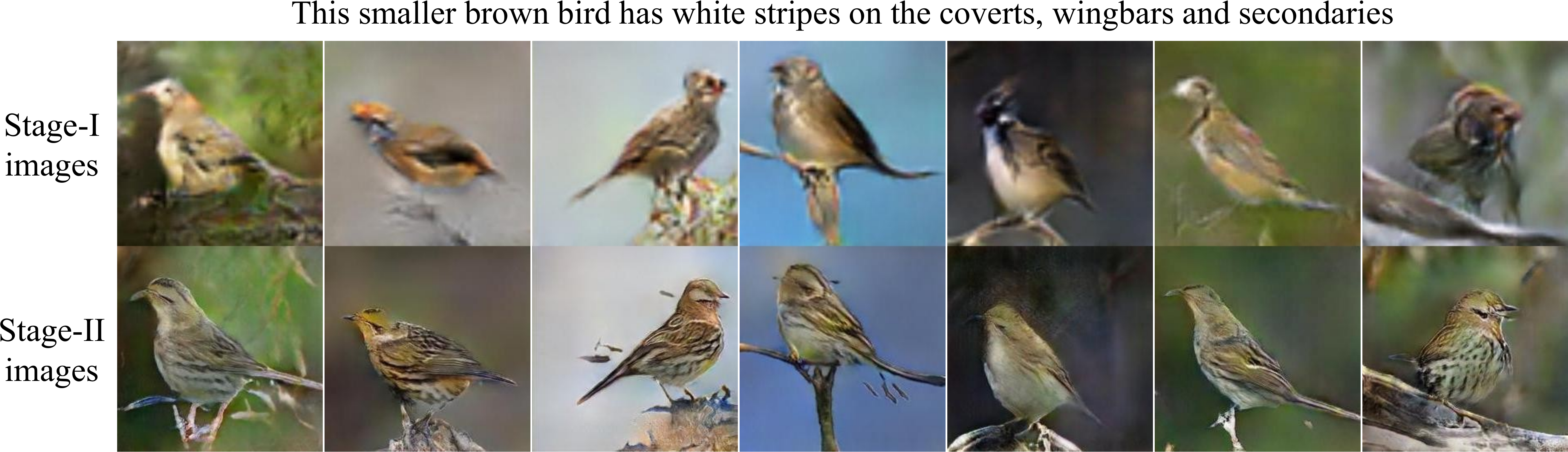}
\vspace{+5pt}

\includegraphics[width=0.95\linewidth]{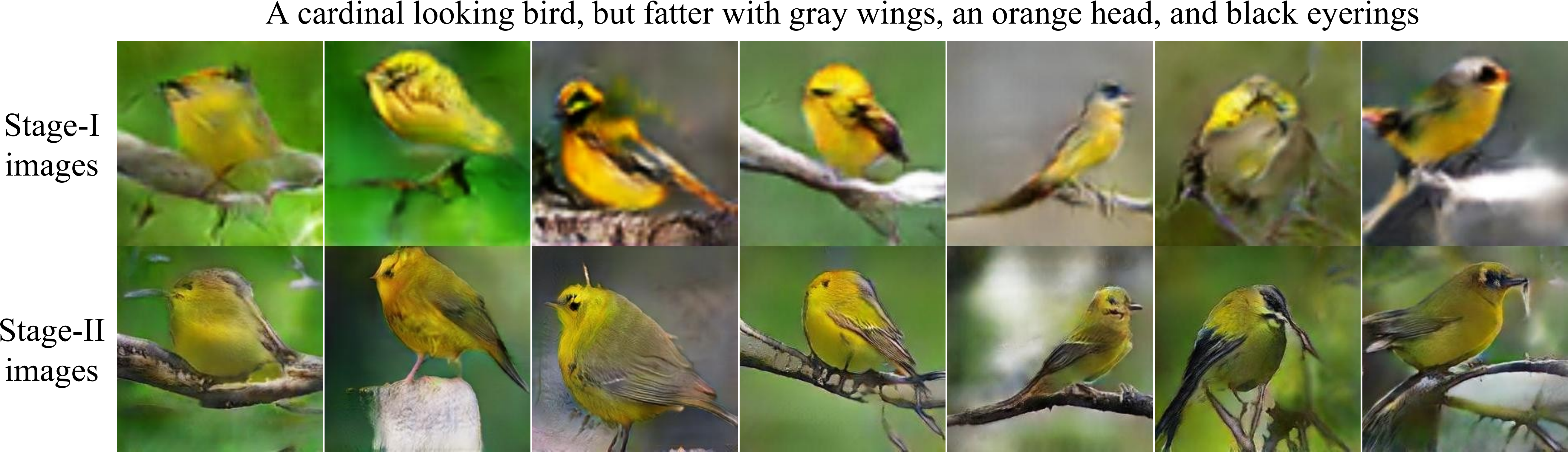}
\vspace{+5pt}

\includegraphics[width=0.95\linewidth]{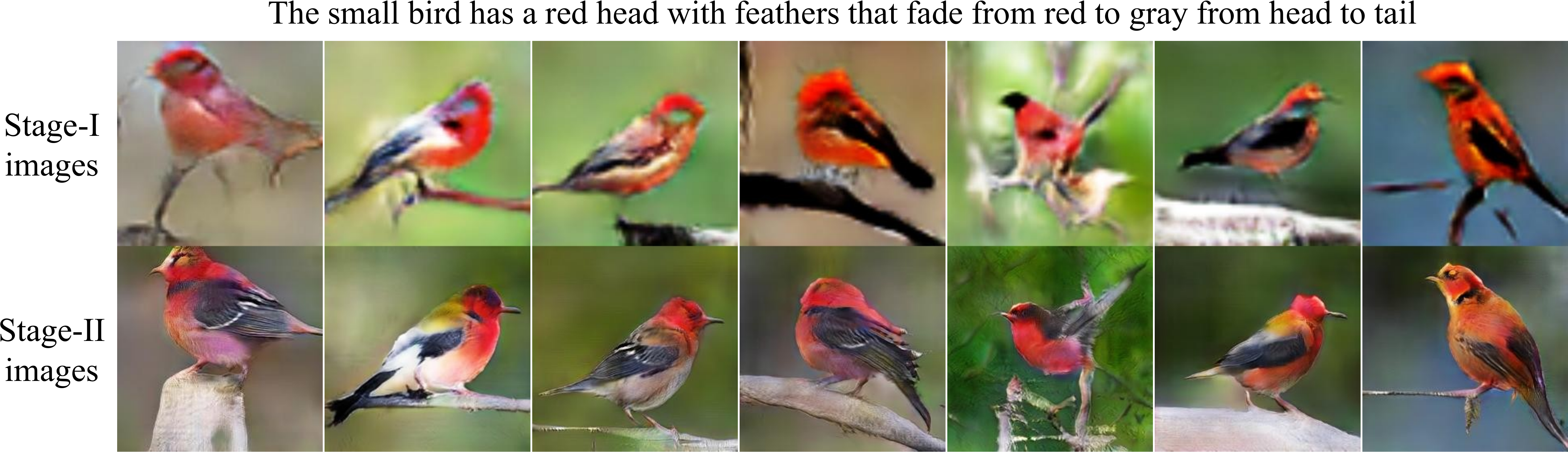}
\vspace{+5pt}

\includegraphics[width=0.95\linewidth]{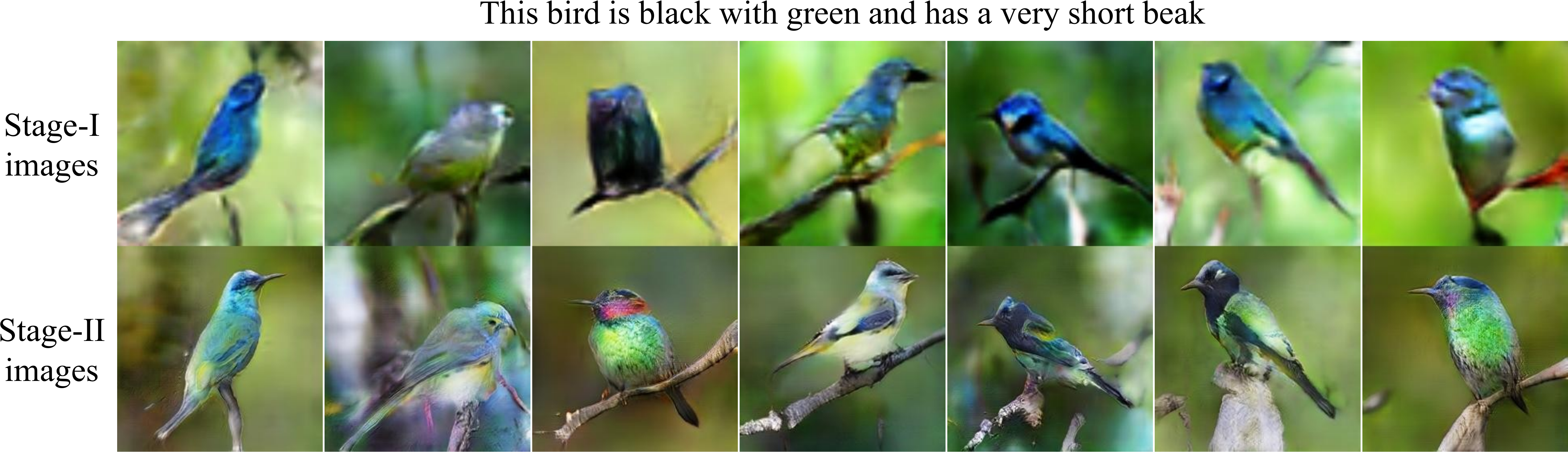}
\vspace{+5pt}


\includegraphics[width=0.95\linewidth]{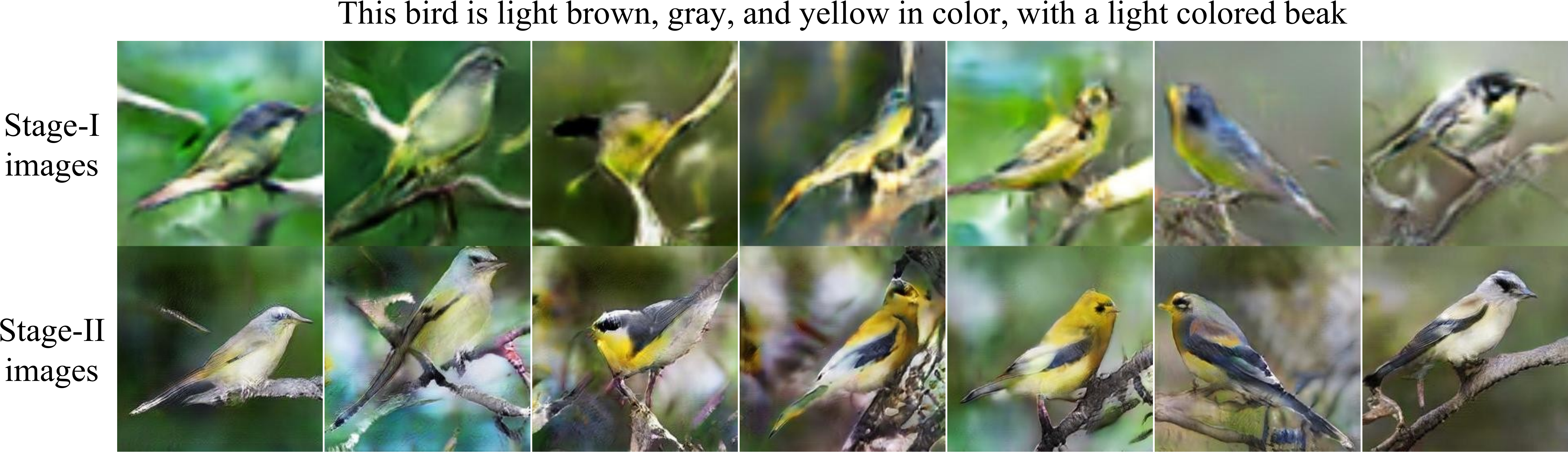}
\vspace{+5pt}

\includegraphics[width=0.95\linewidth]{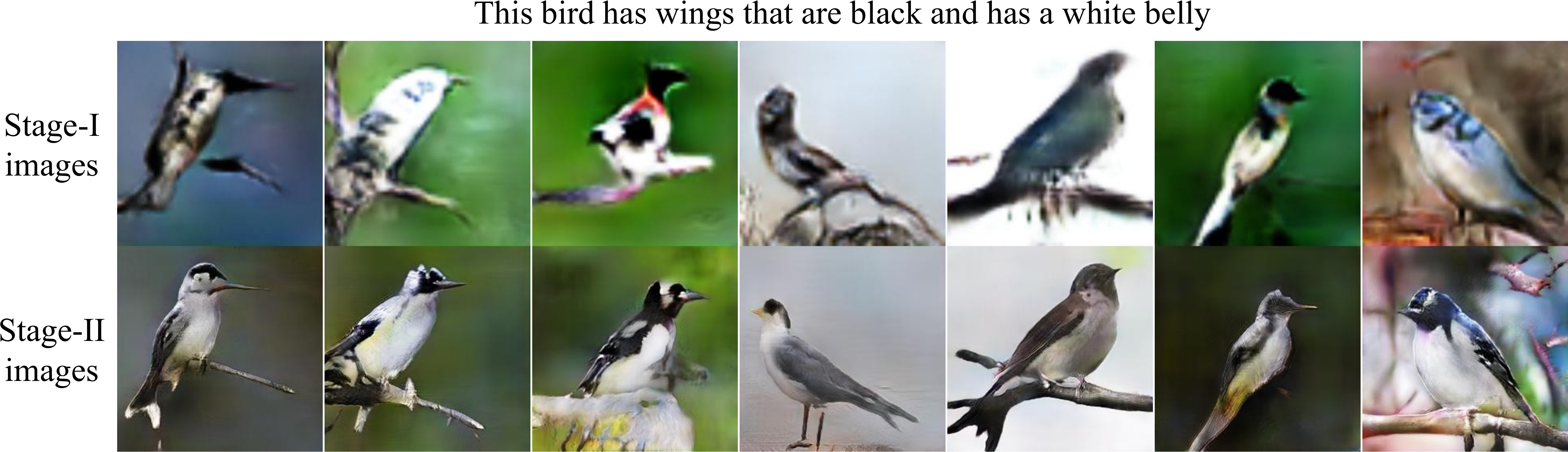}


\clearpage
\textbf{Additional Results on Oxford-102 Dataset}
\vspace{+5pt}

\includegraphics[width=0.95\linewidth]{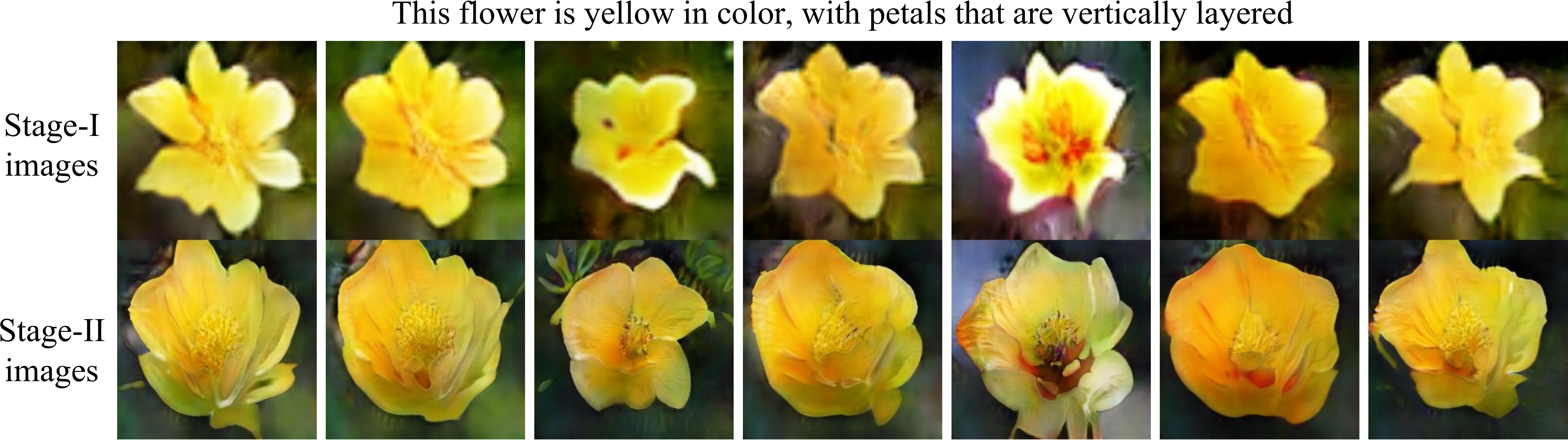}
\vspace{+5pt}

\includegraphics[width=0.95\linewidth]{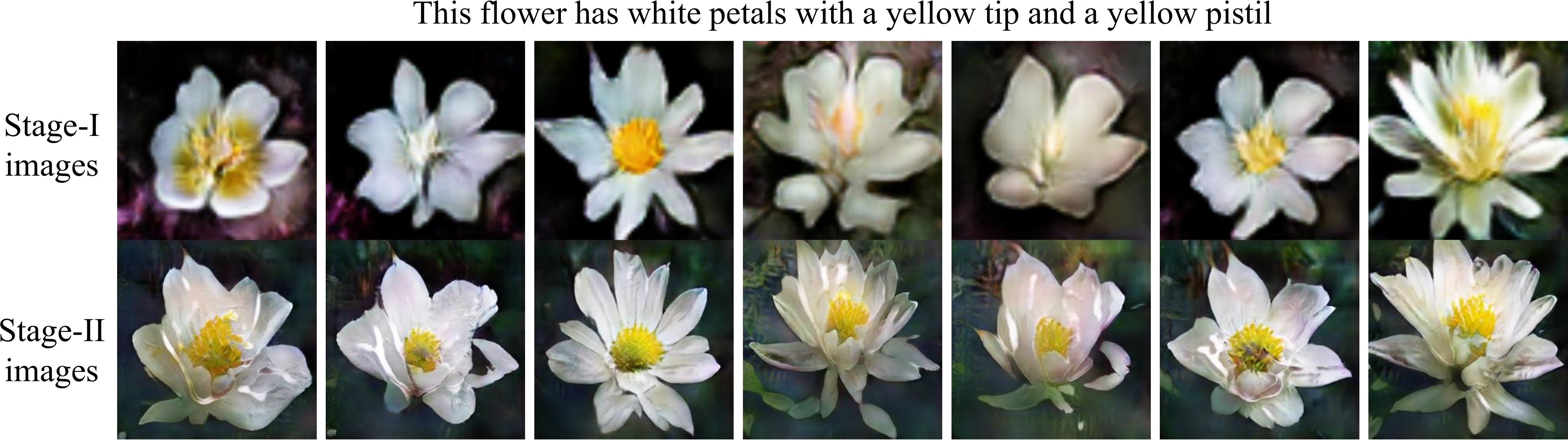}
\vspace{+5pt}

\includegraphics[width=0.95\linewidth]{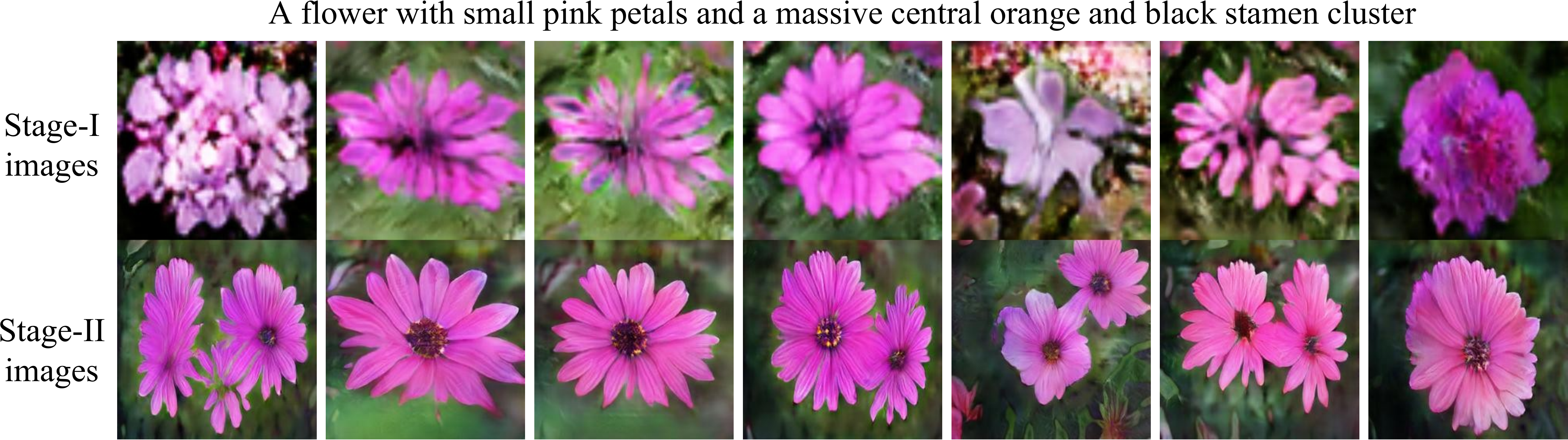}
\vspace{+5pt}

\includegraphics[width=0.95\linewidth]{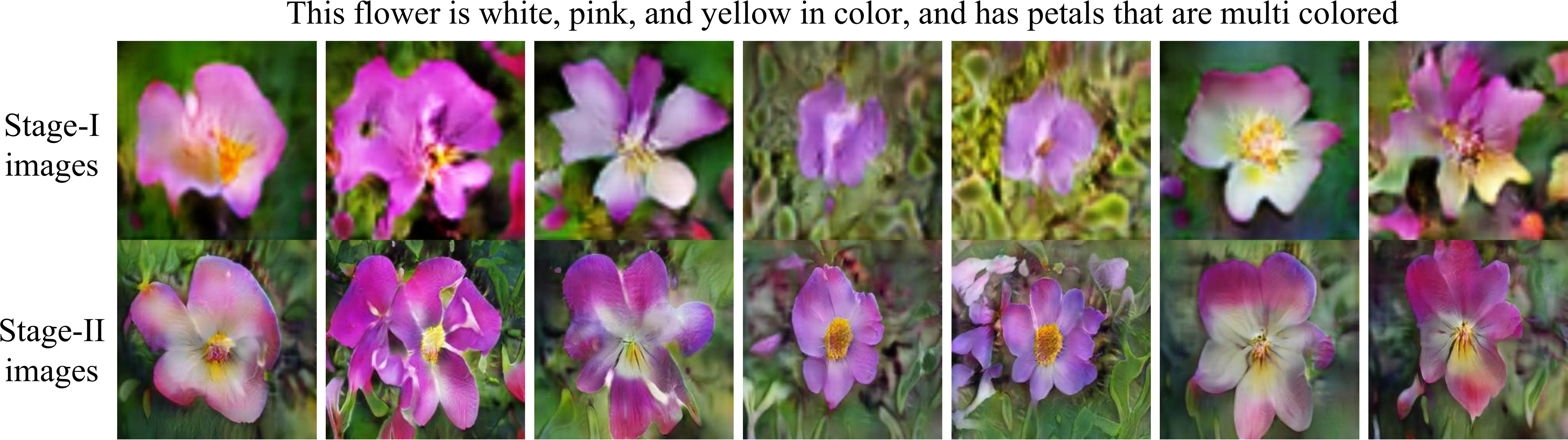}
\vspace{+5pt}

\includegraphics[width=0.95\linewidth]{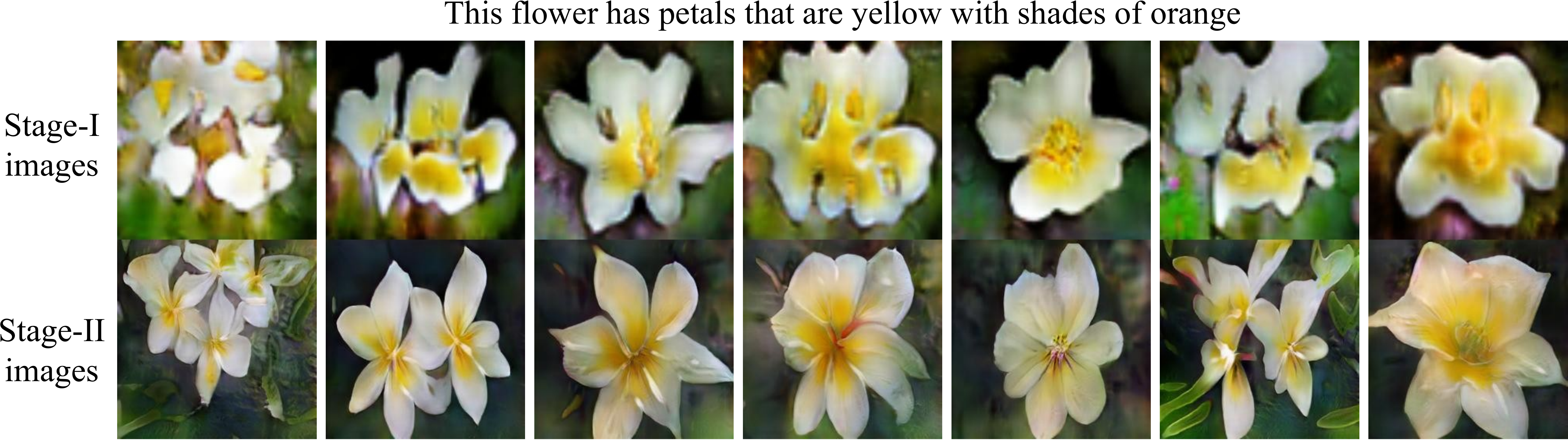}

\subsection*{Failure Cases}
The main reason for failure cases is that Stage-\Rmnum{1} GAN fails to generate plausible rough shapes or colors of the objects. 

\vspace{+5pt}
\textbf{CUB failure cases:}
\vspace{+5pt}

\includegraphics[width=0.95\linewidth]{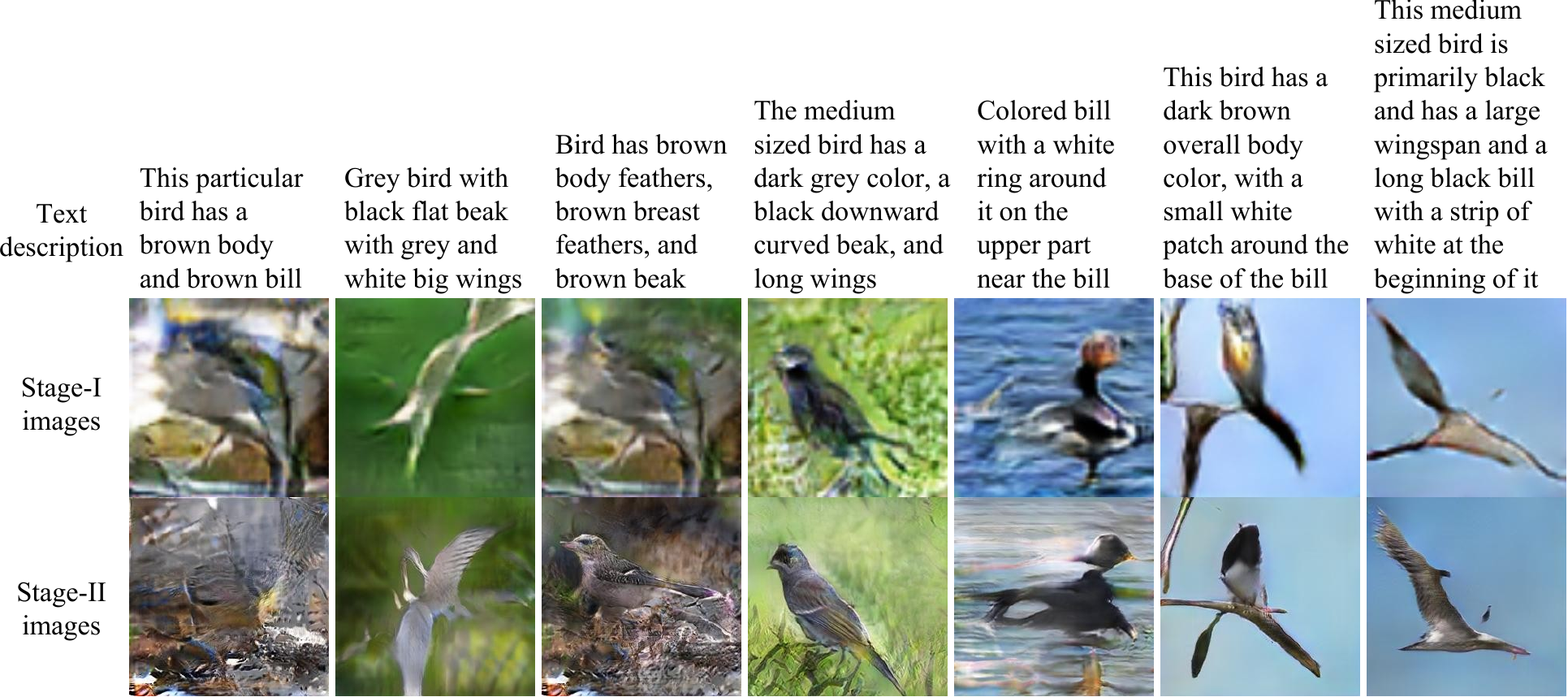}

\vspace{+10pt}
\textbf{Oxford-102 failure cases:}
\vspace{+5pt}

\includegraphics[width=0.95\linewidth]{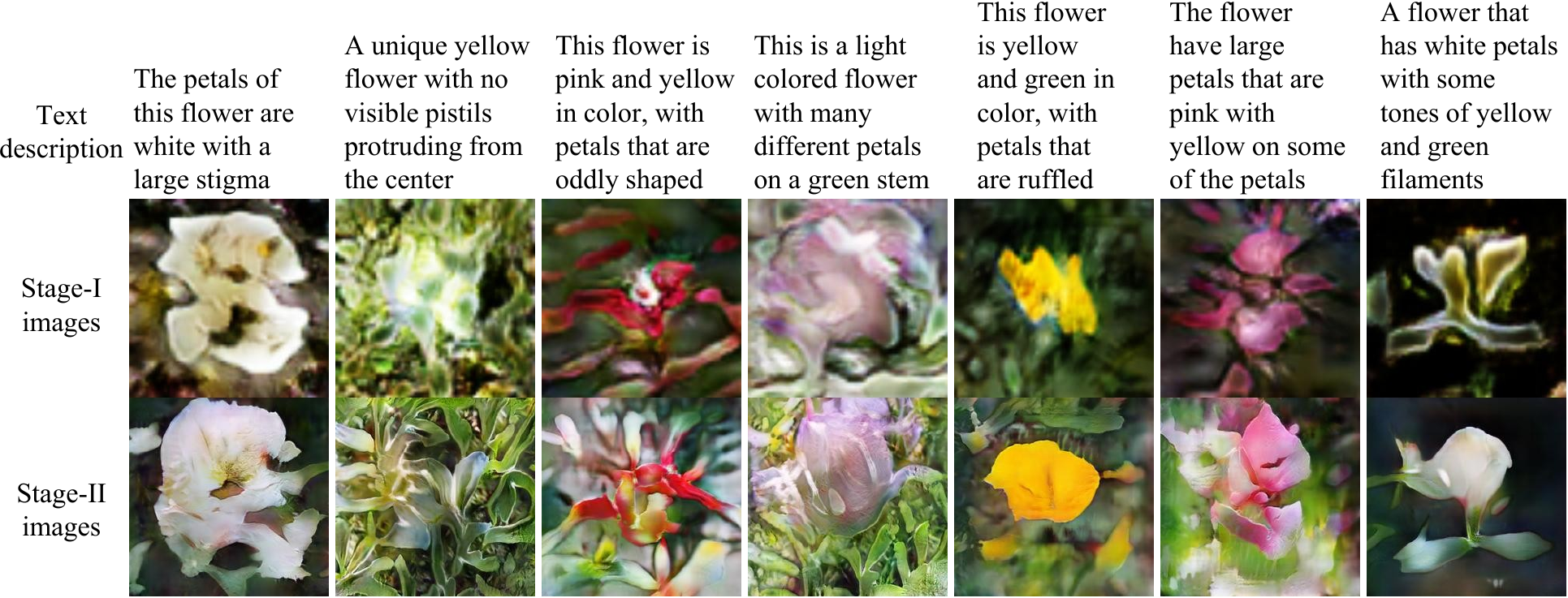}
\vspace{+205pt}

\section*{Beyond Birds and Flowers: Results on MS COCO}
\vspace{+5pt}

Results on COCO dataset demonstrate the generalization capability of our approach on images with multiple objects and complex backgrounds.

\vspace{+5pt}
\textbf{Diverse samples can be generated for each text description. }
\vspace{+5pt}


\includegraphics[width=0.95\linewidth]{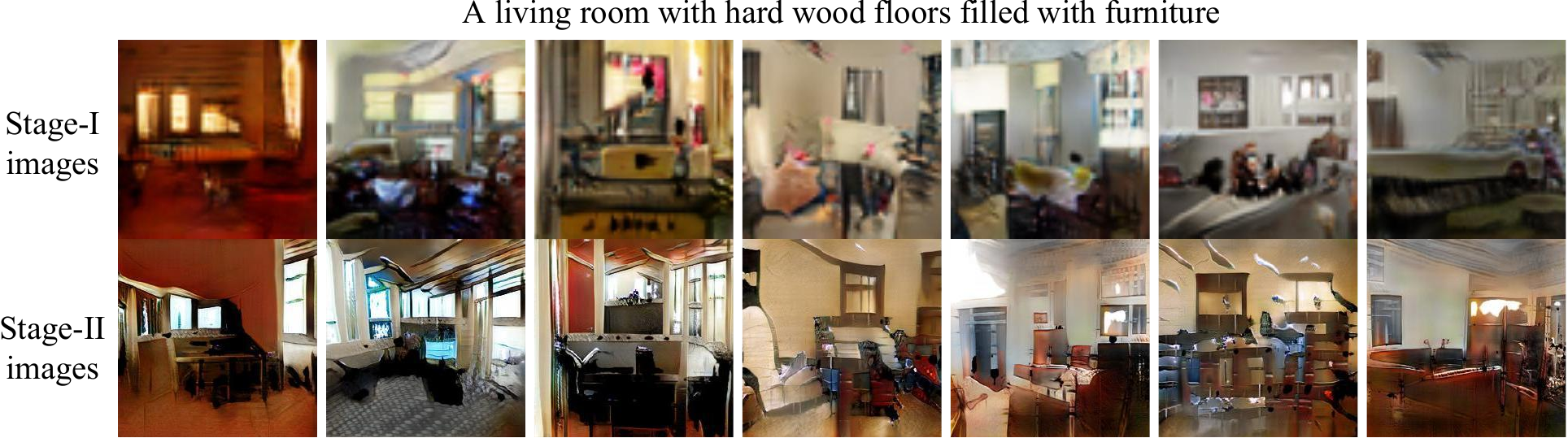}
\vspace{+5pt}

\includegraphics[width=0.95\linewidth]{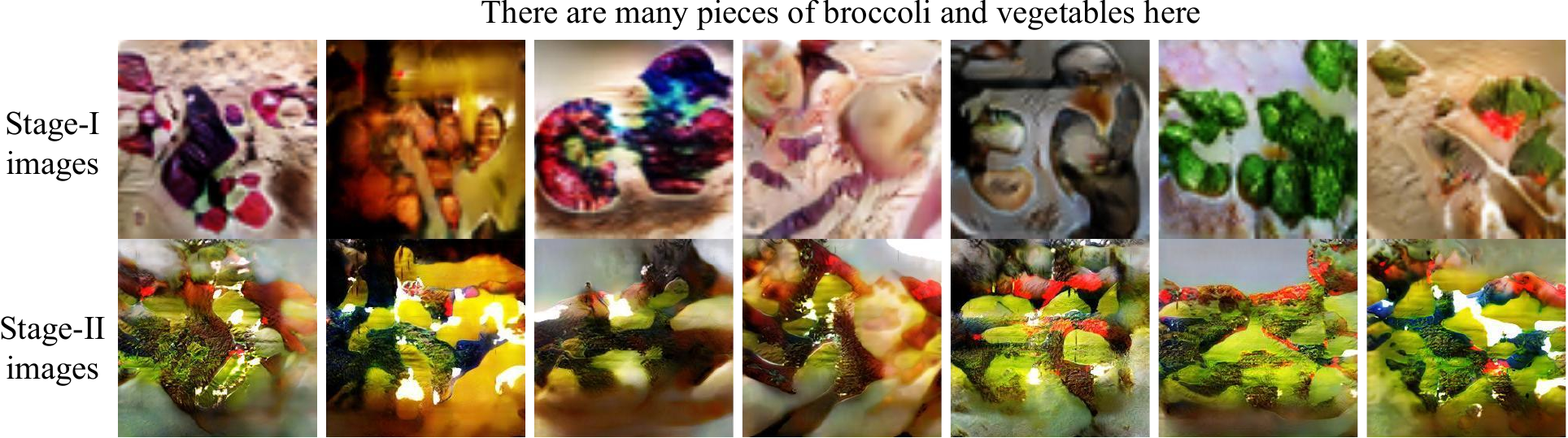}
\vspace{+5pt}

\vspace{+15pt}
\textbf{More results. }
We observe that StackGAN is able to synthesize reasonable images in various cases, although the image quality is lower than the results of birds and flowers. In the future work, we aim to further investigate more sophisticated stacked architectures for generating more complex scenes. 

\vspace{+5pt}
\includegraphics[width=0.95\linewidth]{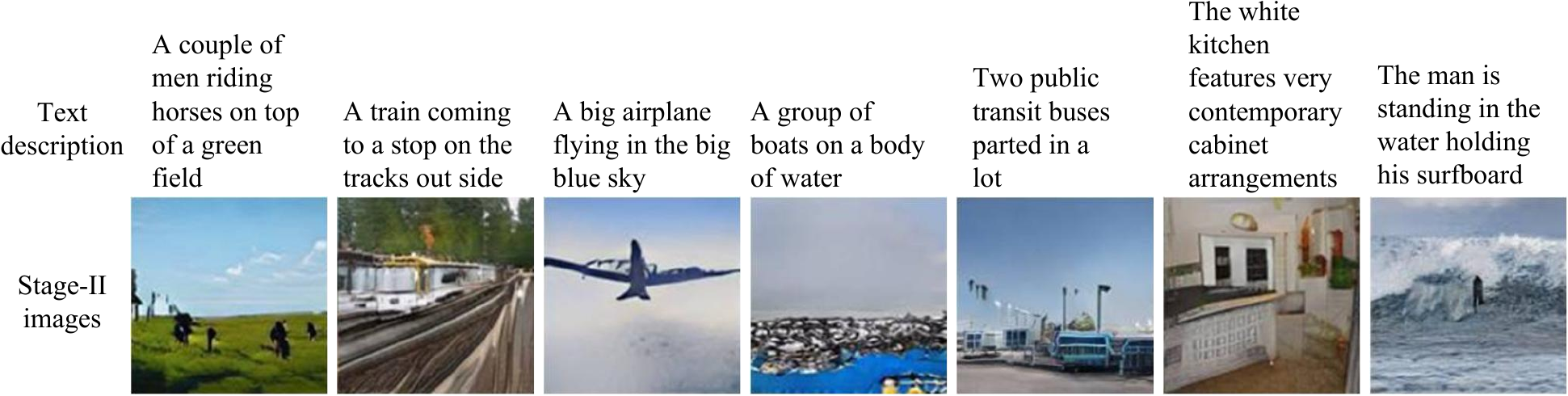}
\vspace{+5pt}

\includegraphics[width=0.95\linewidth]{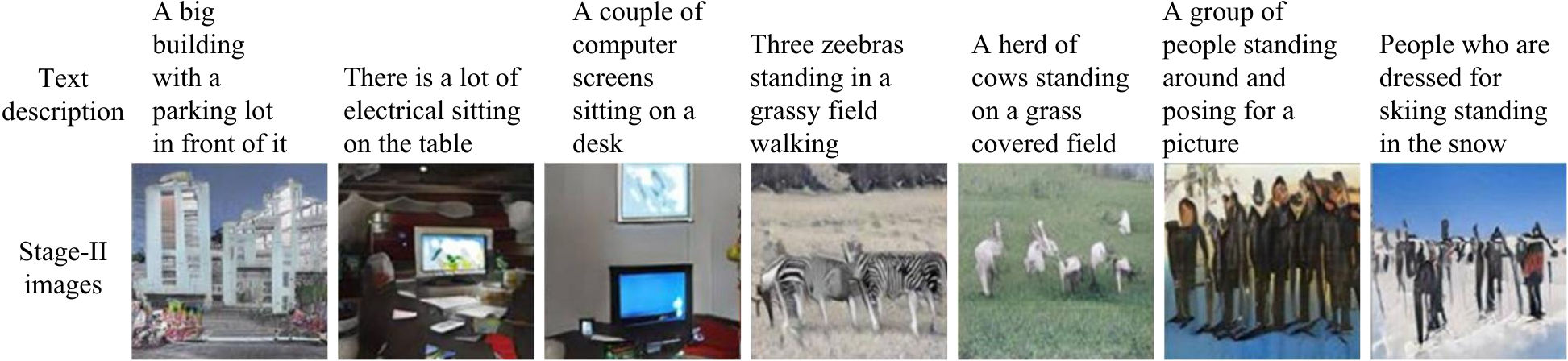}
\end{document}